\documentclass[10pt,journal,compsoc]{IEEEtran}

\usepackage{ifpdf}

\ifCLASSOPTIONcompsoc
  \usepackage[nocompress]{cite}
\else
  \usepackage{cite}
\fi

\ifCLASSINFOpdf
  \usepackage[pdftex]{graphicx}
  \graphicspath{{../pdf/}{../jpeg/}}
  \DeclareGraphicsExtensions{.pdf,.jpeg,.png}
\else
  \usepackage[dvips]{graphicx}
  \graphicspath{{../eps/}}
  \DeclareGraphicsExtensions{.eps}
\fi

\usepackage{amsmath}
\usepackage{array}
\usepackage{algorithm,algpseudocode}

\ifCLASSOPTIONcompsoc
  \usepackage[caption=false,font=footnotesize,labelfont=sf,textfont=sf]{subfig}
\else
  \usepackage[caption=false,font=footnotesize]{subfig}
\fi

\usepackage{dblfloatfix}

\ifCLASSOPTIONcaptionsoff
  \usepackage[nomarkers]{endfloat}
 \let\MYoriglatexcaption\caption
 \renewcommand{\caption}[2][\relax]{\MYoriglatexcaption[#2]{#2}}
\fi

\usepackage{url}
\usepackage{multirow}
\usepackage{booktabs}
\usepackage{bm}
\usepackage{tikz}

\newcommand*\circled[1]{\tikz[baseline=(char.base)]{
            \node[shape=circle,draw,inner sep=0.5pt] (char) {#1};}}

\DeclareMathOperator*{\argmin}{arg\,min}

\newcommand{\STN}{\mathrm{STN}}
\newcommand{\ISTN}{\mathrm{ISTN}}

\newcommand{\bj}{\mathbf{j}}

\newcommand{\pose}{\mathbf{p}}
\newcommand{\calD}{\mathcal{D}}

\newcommand{\calT}{\mathcal{T}}

\DeclareMathOperator{\loc}{loc}
\DeclareMathOperator{\pred}{pred}
\DeclareMathOperator{\synth}{synth}
\DeclareMathOperator{\updater}{updater}
\DeclareMathOperator{\iter}{iter}
\DeclareMathOperator{\proj}{proj}
\newcommand{\inp}{\text{input}}
\usepackage{xspace}

\makeatletter
\DeclareRobustCommand\onedot{\futurelet\@let@token\@onedot}
\def\@onedot{\ifx\@let@token.\else.\null\fi\xspace}

\def\eg{\emph{e.g}\onedot} 
\def\ie{\emph{i.e}\onedot} 
 
\def\etc{\emph{etc}\onedot} 
 
\def\etal{\emph{et al}\onedot}
\makeatother

\begin{document}
%
\title{Generalized Feedback Loop for\\ Joint Hand-Object Pose Estimation}
%
%
%
%

\author{Markus~Oberweger, 
        Paul~Wohlhart, 
        and~Vincent~Lepetit
\IEEEcompsocitemizethanks{\IEEEcompsocthanksitem M.~Oberweger, P.~Wohlhart, and V.~Lepetit are with the Institute for Computer Graphics and Vision, Graz University of Technology, Graz, Austria.\protect\\
E-mail: lastname@icg.tugraz.at
\IEEEcompsocthanksitem V.~Lepetit is also with the Laboratoire Bordelais de Recherche en Informatique, Universit\'{e} de Bordeaux, Bordeaux, France.
\IEEEcompsocthanksitem P.~Wohlhart is now with X, Alphabet Inc., Mountain View, CA.}
\thanks{Manuscript received June 14, 2018; revised November 28, 2018.}}

%
%

\markboth{Journal of \LaTeX\ Class Files,~Vol.~14, No.~8, August~2015}%
{Oberweger \MakeLowercase{\textit{et al.}}: Generalized Feedback Loop}
%



\IEEEtitleabstractindextext{%
\begin{abstract}

We propose an approach to estimating the 3D pose of a hand, possibly handling an
object, given a depth image.  We show that we can correct the mistakes made by a
Convolutional Neural  Network trained to predict  an estimate of the  3D pose by
using  a feedback  loop.  The  components of  this feedback  loop are  also Deep
Networks,   optimized    using   training   data.    This    approach   can   be
generalized  to a  hand  interacting with  an  object.  Therefore,  we
jointly estimate  the 3D pose of  the hand and the  3D pose of the  object.  Our
approach  performs  en-par  with  state-of-the-art  methods  for  3D  hand  pose
estimation, and outperforms state-of-the-art  methods for joint hand-object pose
estimation when using depth images only.  Also, our approach is efficient as our
implementation runs in real-time on a single GPU.
\end{abstract}

}

\maketitle

\IEEEdisplaynontitleabstractindextext

%
\IEEEpeerreviewmaketitle

\IEEEraisesectionheading{\section{Introduction}\label{sec:introduction}}
\IEEEPARstart{A}{ccurate}  hand  pose estimation  is  an  important  requirement for  many  Human
Computer Interaction  or Augmented  Reality tasks~\cite{Erol2007}, and  has been
steadily  regaining ground  as a  focus  of research  interest in  the past  few
years~\cite{Keskin2011,Keskin2012,Melax2013,Oikonomidis2011,Qian2014,Tang2014,Tang2013,Xu2013},
probably because of  the emergence of 3D sensors.  Despite  3D sensors, however,
it is still a  very challenging problem, because of the  vast range of potential
freedoms it  involves, and because  images of hands exhibit  self-similarity and
self-occlusions, and in case of object manipulation occlusions by the object.

A popular approach is to use a discriminative method to predict the position of the joints~\cite{Xu2016,Oberweger2017,Neverova2017,Guo2017,Deng2017,Zhou2016,Tompson2014}, because they are now robust and fast.  To refine the pose further, they are often used to initialize an optimization where a 3D model of the hand is fit to the input depth data~\cite{Ballan2012,LaGorce2011,Oikonomidis2011,Plaenkers03,Qian2014,Sharp2015,Sridhar2013,Tzionas2014}. Such an optimization remains complex and typically requires the maintaining of multiple hypotheses~\cite{Oikonomidis2011a,Oikonomidis2011,Qian2014}.  It also relies on a criterion to evaluate how well the 3D model fits to the input data, and designing such a criterion is not a simple and straightforward task~\cite{Ballan2012,LaGorce2011,Sridhar2013,Tkach2017,Taylor2016}.

In this paper, we first show how we can get rid of the 3D model of the hand altogether and
build instead upon  work that learns to generate images from
training  data~\cite{Dosovitskiy2015}. Creating an anatomically accurate 3D model of the hand is very difficult since the hand contains numerous muscles, soft tissue, \etc, which influence the shape of the hand~\cite{Tagliasacchi2015,Tkach2017,Taylor2016,Sridhar2015}. We think that our approach could also be applied to other problems where acquiring a 3D model is very difficult.

\begin{figure}
\begin{center}
\includegraphics[width=0.9\linewidth]{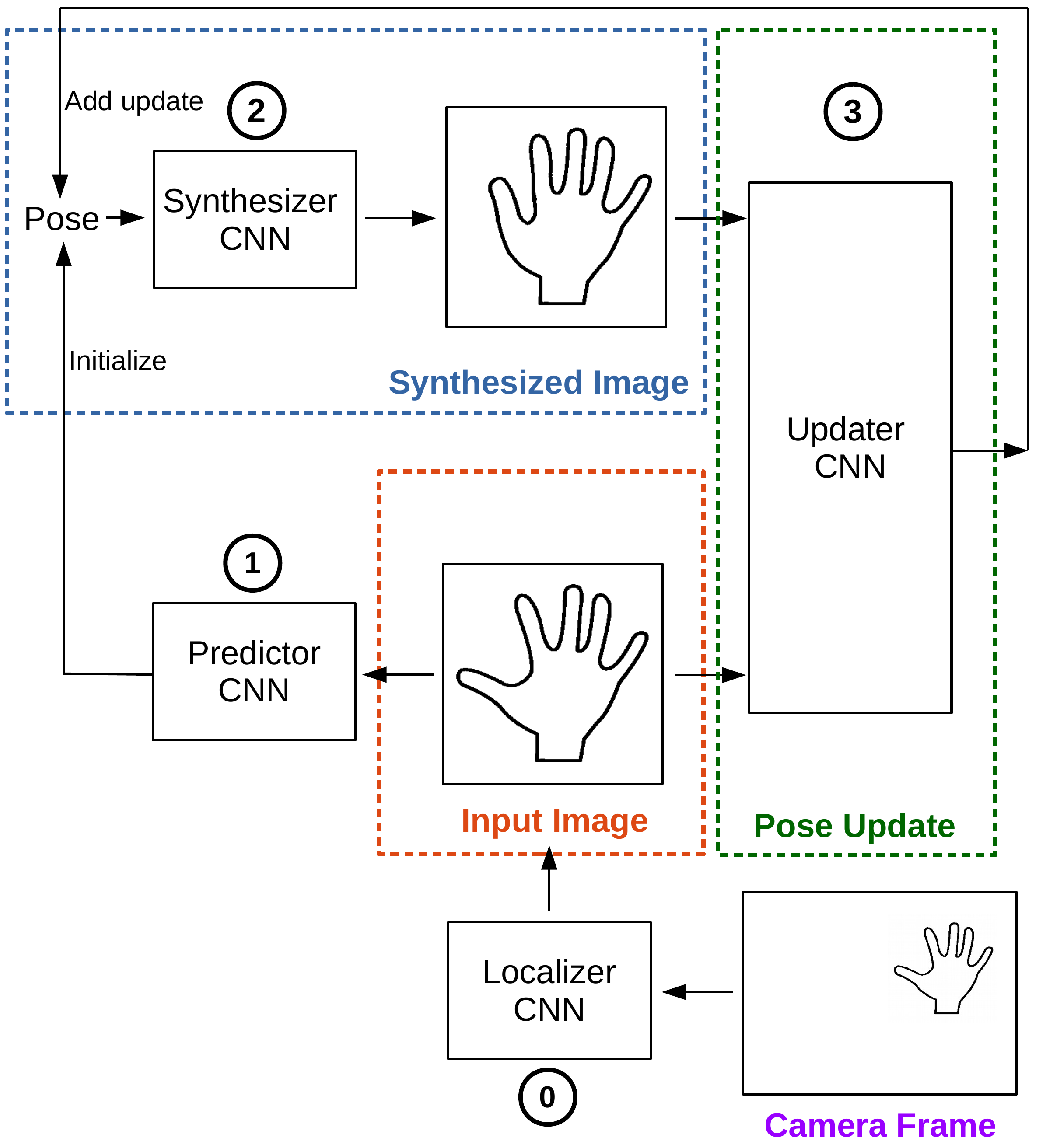}
\end{center}
   \caption{Overview of  our approach for hand pose estimation. First, we localize the hand in the camera frame \protect\circled{0} and crop a patch around the hand location. Then, we  use a first  CNN \protect\circled{1} to predict  an initial
     estimate of the 3D  pose given an input depth image of  the hand.  The pose
     is used to synthesize a depth image \protect\circled{2}, which is used together with the input depth
     image to derive  a pose update \protect\circled{3}. The  update is applied to the  pose and the
     process is iterated.
In this work, we follow this general approach and further show how to extend it to joint hand-object pose estimation.
     }
\label{fig:architecture_overview}
\label{fig:teaser}
\end{figure}

We then  introduce a  method that  learns to provide  updates for  improving the
current estimate of the pose, given the  input depth image and the image generated for
this pose estimate as shown  in Fig.~\ref{fig:teaser}.  By iterating this method
a number of times,  we can correct the mistakes of  an initial estimate provided
by a simple discriminative method. All the components are implemented as Deep Networks with simple architectures.

Not  only is it interesting  to see  that all  the components  needed for  hand
registration that  used to require careful  design can be  learned instead, 
but we will also show that our approach has en-par performance compared to the
state-of-the-art methods. It is also very efficient and runs in real-time on a single GPU.

This method was originally published in~\cite{Oberweger2015a}. Here, we also show how to  generalize our feedback loop to the challenging task of jointly estimating the 3D poses of a hand and an object, while the hand interacts with the object. This is inherently challenging, since the object introduce additional occlusions, and enlarge the joint configuration space. In such a case, we first estimate an initial poses for the hand and the object separately, and then fuse these initial predictions within our feedback framework to increase accuracy of the two poses. For this complex problem, our novel approach works on each frame independently, and does not require a good initialization as current tracking-based approaches do~\cite{Sridhar2016,Tzionas2016}, while still outperforming the state-of-the-art when using depth images only.

Our   approach  is   related   to   generative  approaches~\cite{Bishop06},   in
particular~\cite{Tang2014a} which  also features  a feedback loop  reminiscent of
ours.   However,  our  approach  is   deterministic  and  does  not  require  an
optimization based on  distribution  sampling, on  which generative  approaches
generally rely, but which tends to be inefficient.

In the remainder of  the paper, we first give a short review  of related work in
Section~\ref{sec:relatedwork}.     We    describe    our    approach for hands in isolation in
Section~\ref{sec:main}, introduce the extension for hands and objects in Section~\ref{sec:main_ho}. Finally, we evaluate our method for hand pose estimation in Section~\ref{sec:eval} and for joint hand-object pose estimation in Section~\ref{sec:eval_ho}.

\section{Related Work}
\label{sec:relatedwork}

Hand pose estimation is a frequently visited  problem  in  Computer Vision,  and
is the subject of a plethora of published work. We refer to~\cite{Erol2007} for an overview
of  earlier work, and here we  will discuss  only more  recent work, which can be
divided into two types of approaches.

\subsection{Discriminative Methods} 
The first  type of approaches  is based on discriminative  models that
aim  at directly  predicting the  joint  locations from  RGB or  RGB-D
images. Some recent works, including~\cite{Keskin2011,Keskin2012,Kuznetsova2013,Tang2014,Tang2013} use different approaches with Random Forests, but are restricted to static gestures, showing difficulties with occluded joints, or causing high inaccuracies at finger tips. These problems have been addressed by more recent works~\cite{Oberweger2015,Tompson2014,Mueller2017,Oberweger2017,Zimmermann2017,Xu2016,Neverova2017,Guo2017,Deng2017,Zhou2016} that use Convolutional Neural Networks, nevertheless they are still not as accurate as generative methods that we discuss next.

\subsection{Generative Methods}
Our  approach is  more  related  to  the  second type,  which  covers
generative, model-based methods.  The works of this type are developed  from four generic
building blocks: (1)  a hand model, (2) a similarity  function that measures the
fit  of the  observed image  to  the model,  (3)  an  optimization algorithm  that
maximizes the similarity  function with respect to the  model parameters, and
(4) an initial pose from which the optimization starts.

For the hand model, different hand-crafted models were proposed. The choice of a
simple  model is important for maintaining real-time  capabilities, and representing a
trade-off between  speed and  accuracy as a response  to a potentially  high degree  of model
abstraction. Different  hand-crafted geometrical  approximations for  hand models
were proposed.
For example, \cite{Qian2014}  uses  a  hand model consisting of spheres, 
\cite{Oikonomidis2011} adds cylinders, ellipsoids, and  cones.
\cite{Sridhar2013} models the hand
from  a  Sum   of  Gaussians.  More  holistic  hand   representations  are  used
by~\cite{Ballan2012,Sharp2015,Tzionas2014},  with a  linear blend skinning model
of the hand that is rendered  as depth image. \cite{Xu2013} increases the matching
quality by using depth images that resemble  the same noise pattern as the depth
sensor. \cite{LaGorce2011} uses a fully shaded and textured triangle mesh controlled by a skeleton.

Different  modalities  were proposed for  the  similarity function,  which  are
coupled        to       the        used        model.
The modalities include, \eg, depth values \cite{Melax2013,Oikonomidis2011a,Qian2014}, 
salient points, edges, color \cite{LaGorce2011}, or 
combinations of these \cite{Ballan2012,Oikonomidis2011,Sridhar2013,Tzionas2014}.

The optimization  of the  similarity function  is a critical  part, as  the high
dimensional  pose  space  is  prone  to  local  minima.   Thus,  Particle  Swarm
Optimization  is often  used  to  handle the  high  dimensionality  of the  pose
vector~\cite{Oikonomidis2011a,Oikonomidis2011,Qian2014,Sharp2015}. Differently,
\cite{Ballan2012,LaGorce2011,Sridhar2013} use gradient-based methods
to optimize  the pose, while \cite{Melax2013} uses dynamics simulation.
Due to the high computation time of these optimization methods, which have to be
solved  for every  frame,  \cite{Xu2013} does  not optimize  the  pose but  only
evaluates the similarity function for several proposals to select the best fit.
Recent works aim at handcrafting a differentiable similarity function, which tightly integrates the hand model to achieve accurate and fast results~\cite{Tkach2017,Taylor2016}.

In order to kick-start optimization, \cite{Sridhar2013} uses a
discriminative part-based pose initialization, and \cite{Qian2014}
uses finger  tips only.   \cite{Xu2013} predicts candidate  poses using  a Hough
Forest.   \cite{LaGorce2011} requires  predefined hand  color and  position, and
\cite{Oikonomidis2011}   relies    on   a   manual    initialization.   Furthermore,
tracking-based        methods         use        the pose of the last       
frame~\cite{Melax2013,Oikonomidis2011,Tzionas2014,Sridhar2016},   which  can   be
problematic if the  difference between frames is too  large, due to fast
motion or low frame rates.

Our approach  differs from  previous works in  the first  three building  blocks.
We do not use a deformable CAD model of the hand.  Instead, we learn from
registered depth images to generate realistic  depth images of hands, similar to
work on inverse graphics networks~\cite{Dosovitskiy2015,Kulkarni2015}, and other
recent                   work                    on                   generating
images~\cite{Goodfellow2014,Kulkarni2014}.   This approach  is very
convenient, since deforming  correctly and rendering a  CAD model of the  hand in a realistic manner requires a significant input of engineering work. Most importantly, it does not require additional training data.

In addition, we  do not use  a hand-crafted  similarity function and  an optimization
algorithm.  We learn rather to predict updates that improve
the  current estimate  of the  hand  pose from training data, given  the  input depth  image and  a
generated image for this estimate.  Again,  this approach is very convenient, since it means 
we do not need to design the similarity function and the optimization algorithm,
neither of which is a simple task.

\cite{Nair2008}  relies  on  a  given black-box  image  synthesizer  to  provide
synthetic  samples on  which the  regression network  can be  trained.  It  then
learns  a  network  to  substitute  the  black-box  graphics  model,  which  can
ultimately be used to update the pose  parameters to generate an image that most
resembles the  input.  In contrast, we  learn the generator  model directly from
training data, without the need for a black-box image synthesizer.  Moreover, we
will show that the optimization is prone to output infeasible poses or get stuck
in local minima and therefore introduce a better approach to improve the pose.

\cite{Wan2017} uses an auto-encoder to compute an embedding from the 3D hand pose, and feed the embedding vector to an image synthesizer that outputs a depth image of the hand. This setup is used to synthesize additional training data for training a pose predictor, but in their setup they only aim at generating synthesized images as close as possible to the real images to train a discriminative predictor.

\subsection{Feedback}
Since we  learn to generate images  of the hand, our  approach is also
related to  generative approaches, in particular~\cite{Tang2014a}, which
  uses a  feedback loop with  an updater mechanism akin  to ours.
It predicts updates for the position from which  a patch is
cropped from  an image, such that the patch fits best to the
output of  a generative model. However, this step  does not
predict the  full set of parameters.  The hidden states of  the model
are found by a costly sampling process.

\cite{Carreira2016} proposed an approach related to ours, which also relies on iteratively refining the 2D joint locations for human pose estimation. They use an initial estimate of the 2D joint locations to generate a heatmap with the joint locations. These heatmaps are stacked to the input image and an update is predicted, which, however, does not handle occlusions well. This process is then iterated a few times. In contrast to our work, their approach only works for 2D images due to limitations of the feedback. Also, their training strategy is different, and requires a predefined absolute step size and predefined update directions, which can be hard to learn. In contrast, we use latent update directions with a relative step size.

\cite{Zamir2017} introduced Feedback Networks for general purpose iterative prediction. They formulate the feedback as a recurrent shared operation where a hidden state is passed on to the next iteration. At each iteration, they predict an absolute output and not an update, which is more difficult. Also, their approach does not use any feedback in the input or output space, so their notion of "feedback" is only an internal structure of the predictor.

Since the introduction of our feedback method~\cite{Oberweger2015a}, it was successfully applied to other fields, such as 3D object pose estimation. \cite{Rad2017} predicts updates on 2D object locations. Similarly, \cite{Li2018} uses a CAD rendering of an object together with the input image to predict an update for the rendered CAD pose to better resemble the object pose in the input image.

\subsection{Joint Hand-Object Pose Estimation}

The approaches for hand pose estimation discussed so far can only handle minor occlusion, and in practice the accuracy decreases with the degree of occlusion. To specifically account for occlusions, different approaches try to learn an invariance by using images of a hand with an interacting object for training~\cite{Rogez2015,Mueller2017} to simulate occlusions, by removing the object for estimating the hand pose~\cite{Romero2010}, or by using temporal information to recover from occlusions~\cite{Madadi2017b}.

Similar to hands in isolation, accurate methods are available for object pose estimation without occlusions~\cite{Wohlhart2015,Rad2017}. However, for our problem of joint hand-object pose estimation, object pose methods are required that are specifically robust to partial occlusion. Keypoint-based methods \cite{Lowe2004,Wagner2008} perform well with occlusions, but only on textured objects, and \cite{Crivellaro2015} requires discriminative color and features. In order to handle clutter and occlusions, local patch-based methods~\cite{Brachmann2014,Krull2015}, voting-based methods~\cite{Drost2010}, or segmentation-based methods~\cite{Gupta2015} have shown robustness to clutter and occlusions. However, they only consider objects, and it is not clear how to integrate a hand in these methods.

The problem of hand and object pose estimation is much harder than considering hands or objects in isolation. A straightforward approach of combining methods of both sides does not lead to good results, since the physical constraints and visual occlusions between hand and object are not considered.
Several works considered tracking of two hands simultaneously, however, they require hands to be well-separated~\cite{Wang2011,Tagliasacchi2015}, multiple-views~\cite{Oikonomidis2011}, work at non-interactive framerates~\cite{Oikonomidis2012}, or accurate segmentation~\cite{Hamer2009,Goudie2017}.
For handling unknown objects, \cite{Tzionas2015,Panteleris2015} use an ICP-like object tracking that creates a model of the object on the fly. However, once it looses the temporal tracking, it fails completely.
\cite{Tzionas2016} uses a model-based approach for tracking hand-hand and hand-object poses. Similarly, \cite{Sridhar2016} performs model-based tracking of hand and object jointly, but requires a model for the hand and the object. 
Similar tracking-based approaches where proposed by \cite{Kyriazis2014}, who consider tracking multiple hands and objects jointly at non-interactive framerates, and by \cite{Oikonomidis2011}, who track a single hand interacting with an object but require several cameras for handling the occlusions.
These tracking-based approaches require a good initialization for each frame and are inherently prone to drifting over time. In contrast to these hand-object pose estimation methods, our approach only requires a single depth camera, works without initialization, without temporal tracking, and does not drift over time since we process each frame independently.

\section{Method}
\label{sec:main}

In this  section, we will  first give an  overview of our  method.  We
then  describe in  detail the  different components  of our  method: A
discriminative approach to predict a first pose estimate, a method
able to  generate realistic depth  images of  the hand, and  a learning-based
method to refine the initial pose estimate using the generated depth images.

\subsection{Hand Pose Estimation}
\label{sec:overview}

Our objective is  to estimate the pose $\pose$  of a hand in the form  of the 3D
locations of  its joints $\pose =  \{\bj_i\}^J_{i=1}$ with $\bj_i=(x_i,y_i,z_i)$
from a single depth  image $\calD$. In practice, $J=14$ for  the dataset we use.
We assume that a training set  $\calT = \{(\calD_i, \pose_i)\}^N_{i=1}$ of depth
images labeled with the corresponding 3D joint locations is available.

As explained in the  introduction, we first train a predictor  CNN to predict an
initial pose estimate $\widehat\pose^{(0)}$ in  a discriminative manner given an
input depth image $\calD_\inp$ centered around the hand location:
\begin{equation}
\widehat\pose^{(0)} = \pred(\calD_\inp) \;\; .
\end{equation}
We use a Convolutional Neural Network (CNN) to implement the $\pred(.)$ function
with   a    standard   architecture.   More    details   will   be    given   in
Section~\ref{sec:pred}.

\begin{figure}
\begin{center}
\includegraphics[width=0.24\linewidth,clip,trim={0.5cm 0.5cm 0cm 0cm}]{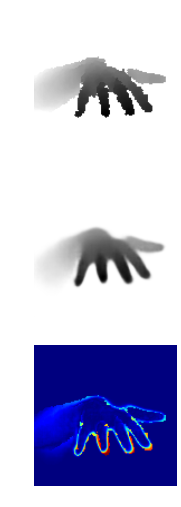}
\includegraphics[width=0.24\linewidth,clip,trim={0.5cm 0.5cm 0cm 0cm}]{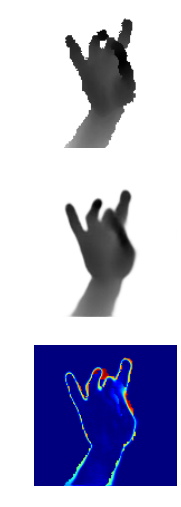}
\includegraphics[width=0.24\linewidth,clip,trim={0.5cm 0.5cm 0cm 0cm}]{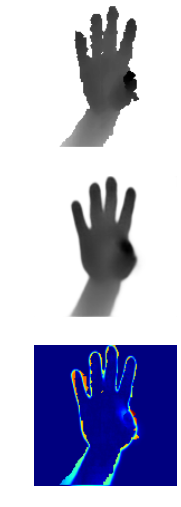}
\includegraphics[width=0.24\linewidth,clip,trim={0.5cm 0.5cm 0cm 0cm}]{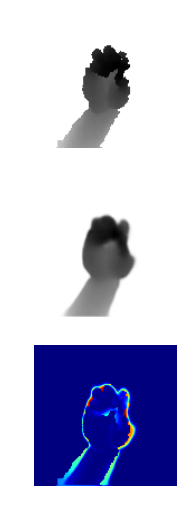}
\end{center}
   \caption{Samples generated  by the synthesizer CNN for different poses  from the
     test  set.  \textbf{Top:} Ground truth depth  image. \textbf{Middle:}
     Synthesized  depth image  using our  learned hand  model.  \textbf{Bottom:}
     Color-coded, pixel-wise difference between  the depth images. Red represents large errors, blue represents small errors.  The synthesizer CNN
     is able to render convincing depth images  for a very large range of poses.
     The largest  errors are located  near the  occluding contours of  the hand,
     which are noisy in the ground truth images.}
\label{fig:gensamples}
\end{figure}

In practice,  $\widehat\pose^{(0)}$ is  never perfect, and  following the motivation provided  in the
introduction, we  introduce a  hand model  learned from  the training  data.  As
shown in  Fig.~\ref{fig:gensamples}, this model  can synthesize the  depth image
corresponding to a  given pose $\pose$, and  we will refer to this  model as the
\emph{synthesizer CNN}:
\begin{equation}
\calD_{\synth} = \synth(\pose) \;\; .
\end{equation}
We also use a Deep Network to implement the synthesizer.

\begin{figure}
\begin{center}
\subfloat[]{\includegraphics[width=0.24\linewidth,clip,trim={1cm 1cm 1cm 1cm}]{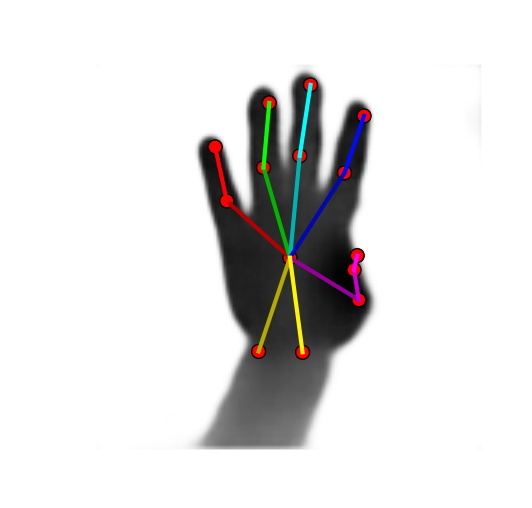}}
\subfloat[]{\includegraphics[width=0.24\linewidth,clip,trim={1cm 1cm 1cm 1cm}]{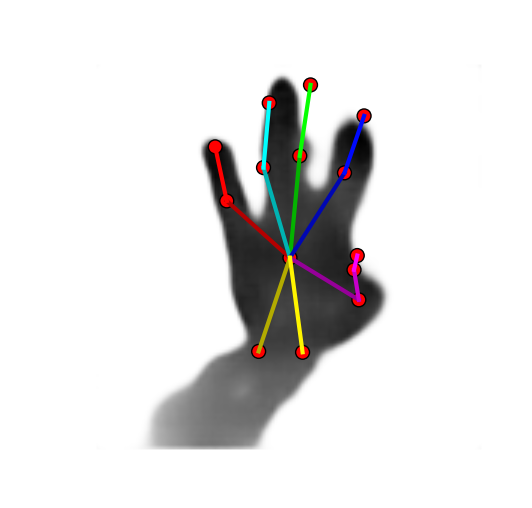}}
\subfloat[]{\includegraphics[width=0.24\linewidth,clip,trim={1cm 1cm 1cm 1cm}]{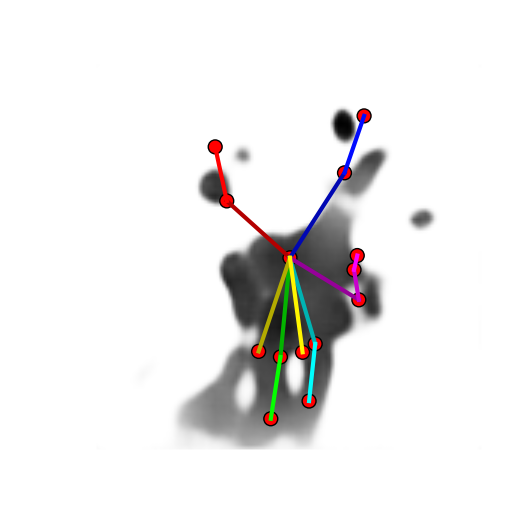}}
\subfloat[]{\includegraphics[width=0.24\linewidth,clip,trim={1cm 1cm 1cm 1cm}]{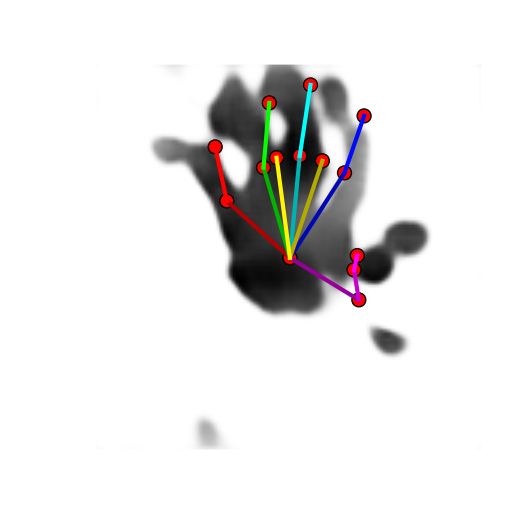}}
\end{center}
   \caption{Synthesized images  for physically impossible  poses. Note
     the colors that indicate different  fingers. (a) shows a feasible
     pose with its synthesized image.  (b) shows the synthesized image
     for the same pose after swapping the  ring and  middle finger positions. In  (c) the
     ring  and middle  finger are  flipped downwards,  and in  (d) the
     wrist joints are flipped upwards.}
\label{fig:gensamples_impossible}
\end{figure}

A straightforward way of using this synthesizer would consist in estimating the hand pose
$\widehat\pose$ by minimizing the squared loss  between the input image and the
synthetic one: 
\begin{equation}
\widehat\pose = \argmin_\pose \lVert \calD_\inp - \synth(\pose) \rVert^2 \;\; .
\label{eq:naive}
\end{equation}
This  is a  non-linear least-squares  problem, which  can potentially be  solved iteratively
using  $\widehat\pose^{(0)}$  as  initial   estimate.   However,  the  objective
function of  Eq.~\eqref{eq:naive} exhibits many local  minima.  Moreover, during
the  optimization   of  Eq.~\eqref{eq:naive},  $\pose$  can   take  values  that
correspond  to  physically  infeasible  poses.   For  such  values,  the  output
$\synth(\pose)$   of   the  synthesizer CNN  is   unpredictable   as  depicted   in
Fig.~\ref{fig:gensamples_impossible}, and this is likely to make the optimization of
Eq.~\eqref{eq:naive} diverge or  be stuck in a local minimum,  as we will show
in the experiments in Section~\ref{sec:naive}.

We  therefore introduce  a third  function that we call the $\updater(.,.)$. It learns to predict updates, which are applied to the pose estimate to improve it, given the input image $\calD_\inp$ and the image $\synth(\pose)$ produced by the synthesizer CNN:
\begin{equation}
\widehat\pose^{(i+1)} \;\leftarrow\; \widehat\pose^{(i)} + \updater(\calD_\inp, \synth(\widehat\pose^{(i)})) \;\; .
\label{eq:updater}
\end{equation}
We   iterate  this   update  several   times   to  improve   the  initial   pose
estimate. Again, the $\updater(.,.)$ function is implemented as a Deep Network.

We detail below  how we implement and train the $\pred(.)$,  $\synth(.)$, and $\updater(.,.)$
functions.

\subsection{Learning the Localizer Function $\loc(.)$}
\label{sec:loc}

\begin{figure}[t]
\begin{center}
\includegraphics[width=0.95\linewidth]{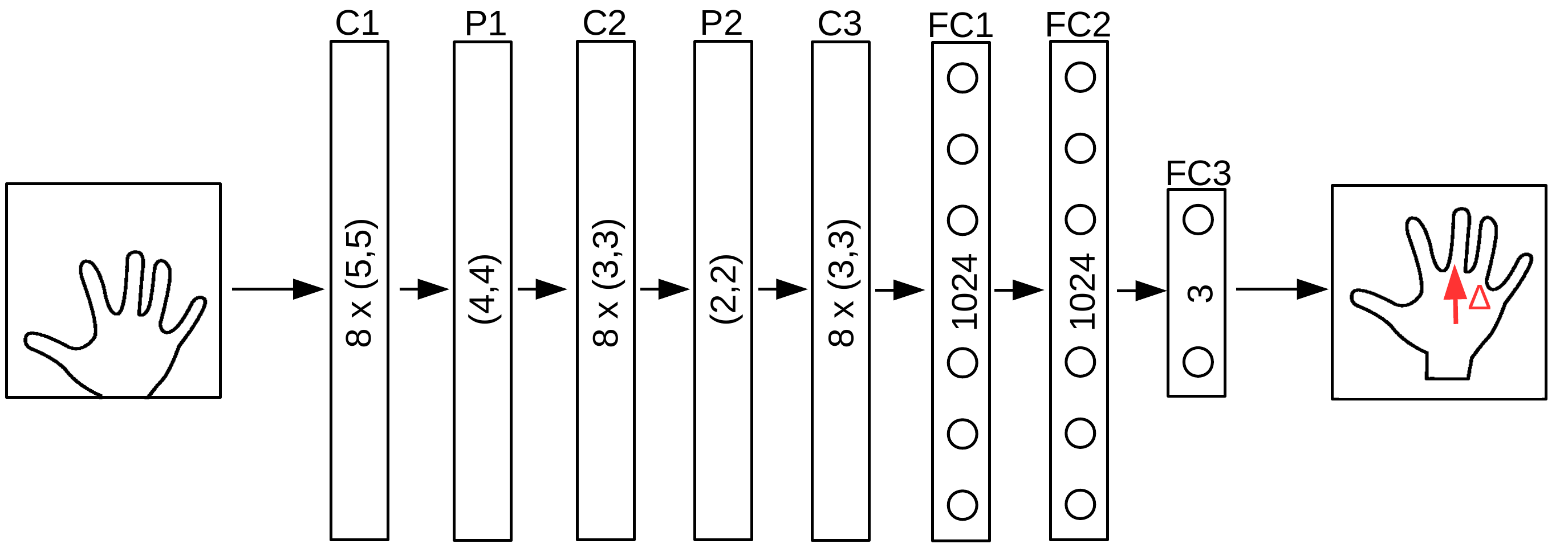}
\end{center}
   \caption{Network  architecture  of the  localizer CNN. All layers have rectified-linear units, except the last layer which has linear units.  \textsf{C}  denotes a
     convolutional  layer  with  the  number  of filters  and  the  filter  size
     inscribed, \textsf{FC} a fully connected  layer with the number of neurons,
     and \textsf{P} a max-pooling layer with the window size. The initial hand crop from the depth image
is fed to the network that predicts the location of the MCP of the hand to correct an
inaccurate hand localization.}
\label{fig:architecture_loc}
\end{figure}

In practice, we require an input depth image centered on the hand location. In our original paper~\cite{Oberweger2015a}, we relied on a common heuristic that uses the center-of-mass for localizing the hand~\cite{Oberweger2015,Tang2014}. In this work, we improve this heuristic by introducing the localizer CNN $\loc(.)$. We still use the center-of-mass for the initial localization from the depth camera frame, but apply an additional refinement step that improves the final accuracy~\cite{Mueller2017,Oberweger2017}. This refinement step relies on the localizer CNN. The CNN is applied to the 3D bounding box centered on the center-of-mass, and is trained to predict the location of the Metacarpophalangeal (MCP) joint of the middle finger, which we use as referential. The localizer CNN has a simple network architecture, which is shown in Fig.~\ref{fig:architecture_loc}.

We train the localizer CNN, parametrized by $\phi$,  by optimizing the cost function:
\begin{equation}
\label{eq:train_loc}
\widehat\phi = \argmin_{\phi} \sum_{(\calD, \pose) \in \calT} \lVert \loc_\phi(\calD) - \mathbf{l} \rVert^2 \;\; .
\end{equation}
$\mathbf{l}$ denotes the offset in image coordinates and depth between the center-of-mass and the MCP of the hand.
For inference, we crop a depth image from the depth camera frame centered on the center-of-mass, then predict the MCP location by applying $\loc(.)$ to this crop, and finally crop again from the depth camera frame using the predicted location.

We crop the training depth  images and the original input depth images
around the locations  provided by $\loc_\phi(.)$ for these  images: $\calD$ in
Eqs.~\eqref{eq:train_pred},  \eqref{eq:train_synth}, \eqref{eq:train_refine2},
\eqref{eq:train_update_poses},                      \eqref{eq:train_jointly1},
\eqref{eq:train_jointly2}  therefore  denotes  a training  depth  image  after
cropping, and $\calD_\inp$ the original input depth image after cropping.

\subsection{Learning the Predictor Function $\pred(.)$}
\label{sec:pred}

\begin{figure}
\begin{center}
   \includegraphics[width=0.95\linewidth]{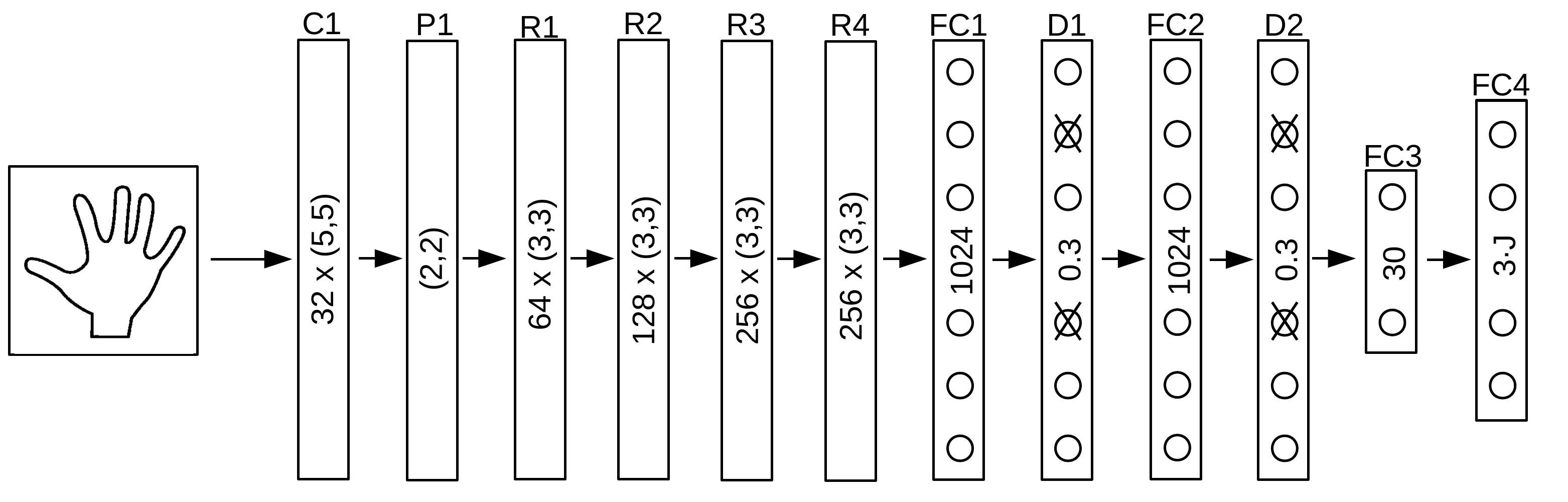}
\end{center}
   \caption{Network  architecture  of the  predictor CNN used for initial pose prediction. \textsf{C} denotes a convolutional
     layer with the  number of filters and the  filter size inscribed,
     \textsf{FC} a  fully-connected layer with the  number of neurons,
     \textsf{D} a  Dropout layer~\cite{Srivastava2014}  with the  probability of  dropping a
     neuron, \textsf{R} a  residual module~\cite{He2016} with the number  of filters and
     filter size, and \textsf{P} a  max-pooling layer with the window
     size. The cropped depth image is fed to the ResNet
     that predicts the 3D hand pose. The last layer is a bottleneck layer with 30 neurons that incorporates the pose prior.~\cite{Oberweger2017}}
\label{fig:resnet}
\end{figure}

The predictor CNN $\pred(.)$ is implemented as a neural network. For our experiments with hands in isolation, we use DeepPrior++~\cite{Oberweger2017} as initialization with the architecture shown in Fig.~\eqref{fig:resnet}.  In the spirit of~\cite{Oberweger2015,Oberweger2017}, we use a prior on the 3D hand pose, which is integrated into the predictor CNN of the hand. We rely on a simple linear prior $\mathbf{P}(.)$, obtained by a PCA of the 3D hand poses~\cite{Oberweger2015}. Thus, $\pred(.)$ predicts the parameters of the pose prior instead of the 3D joint locations, which are obtained by applying the inverse prior to the estimated parameters.  The predictor CNN is  parametrized by
$\Phi$, which is obtained by minimizing

\begin{equation}
\label{eq:train_pred}
\widehat\Phi = \argmin_{\Phi} \sum_{(\calD, \pose) \in \calT} \lVert \mathbf{P}^{-1}(\pred_\Phi(\calD)) - \pose \rVert^2 \;\; .
\end{equation}

\subsection{Learning the Synthesizer Function $\synth(.)$}
\label{sec:synth}

\begin{figure}[t]
\begin{center}
\includegraphics[width=0.95\linewidth]{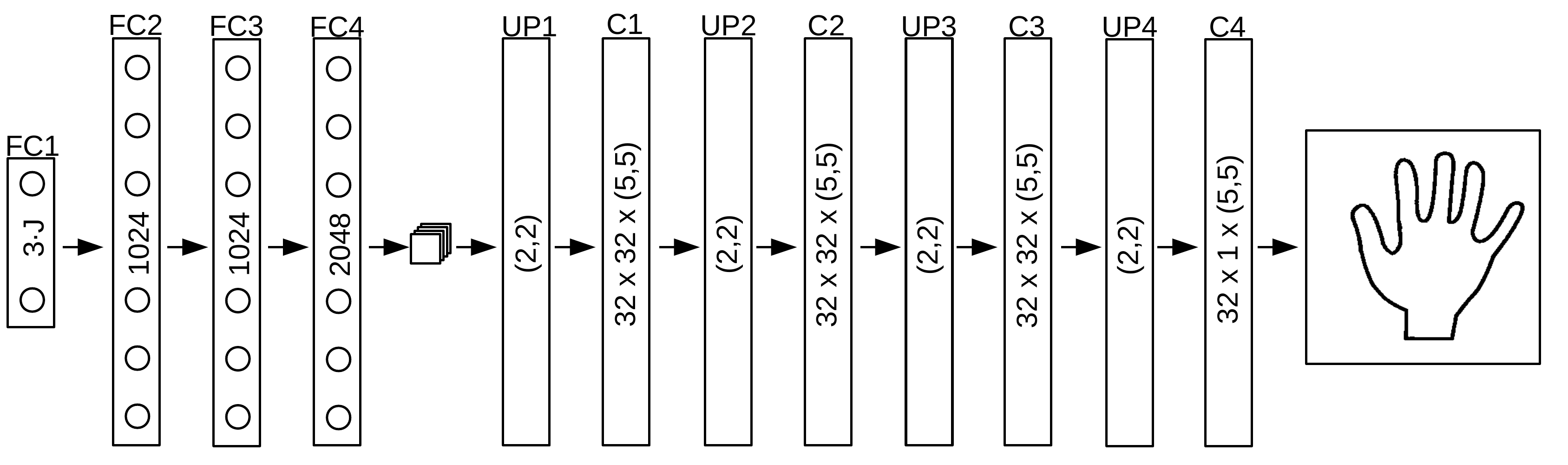}
\end{center}
   \caption{Network  architecture  of the  synthesizer CNN used  to generate  depth
     images of  hands given their  poses. The input of  the network is  the hand
     pose.  The fully connected hidden layers create a 2048 dimensional latent representation 
     at \textsf{FC4} which is reshaped into 32 feature  maps of size  $8\times 8$.
     The  feature  maps  are gradually  enlarged  by  successive
     unpooling and convolution operations.   The last convolution layer combines
     the feature  maps to  derive a  single depth  image of size $128 \times 128$. All layers have rectified-linear units, except the last layer which has tanh units.  \textsf{C}  denotes a
     convolutional  layer  with  the  number  of filters  and  the  filter  size
     inscribed, \textsf{FC} a fully connected  layer with the number of neurons,
     and \textsf{UP} an unpooling layer with the upscaling factor.}
\label{fig:architecture_synth}
\end{figure}

We also use a neural network to implement the synthesizer CNN $\synth(.)$, and we train it using the set $\calT$
of  annotated training  pairs.  The  network architecture  is strongly  inspired
by~\cite{Dosovitskiy2015},  and  is  shown in  Fig.~\ref{fig:architecture_synth}.   It
consists of four hidden layers, which  learn an initial latent representation of
feature  maps apparent  after  the fourth  fully  connected layer  \textsf{FC4}.
These  latent feature  maps are  followed by  several unpooling  and convolution
layers.       The      unpooling      operation,      used      for      example
by~\cite{Dosovitskiy2015,Goodfellow2014,Zeiler2011},  is the  inverse
of the  max-pooling operation:  The feature map  is expanded, in  our case  by a
factor of  2 along each image  dimension.  The emerging "holes"  are filled with
zeros.  The expanded feature maps are  then convolved with trained 3D filters to
generate another set of feature maps. These unpooling and convolution operations
are applied subsequently.  The last  convolution layer combines all feature maps
to generate a depth image.

We learn the parameters $\widehat\Theta$  of the network by minimizing
the   difference   between   the   generated   depth   images
$\synth(\pose)$ and the training depth images $\calD$ as
\begin{equation}
\label{eq:train_synth}
\widehat\Theta = \argmin_{\Theta} \sum_{(\calD, \pose) \in \calT} \frac{1}{|\calD|} \lVert \synth_\Theta(\pose) - \calD \rVert^2 \;\; .
\end{equation}
We perform  the  optimization  in  a  layer-wise fashion. We  start  by training
the feature maps of resolution $8\times 8$. Then we gradually enlarge   the
output  resolution by  adding another  unpooling and  convolutional  layer  and  train
again, which achieves  lesser errors than  end-to-end training  in our
experience.

The synthesizer CNN is able to generate accurate images, maybe surprisingly well for
such a  simple architecture.  The median pixel  error on the  test set  of the NYU dataset~\cite{Tompson2014}  is only
0.1~mm. However, the average pixel error is $8.9$~mm with a standard deviation of $28.5$~mm.  This is mostly due
to noise in  the input images along  the outline of the hand,  which is smoothed
away  by the  synthesizer CNN.  The  average depth  accuracy of  the sensor  is 
$11$~mm~\cite{Khoshelham2012} in practice.

\subsection{Learning the Updater Function $\updater(.,.)$}
\label{sec:feedback}
\begin{figure}[tb]
\begin{center}
\includegraphics[width=0.95\linewidth]{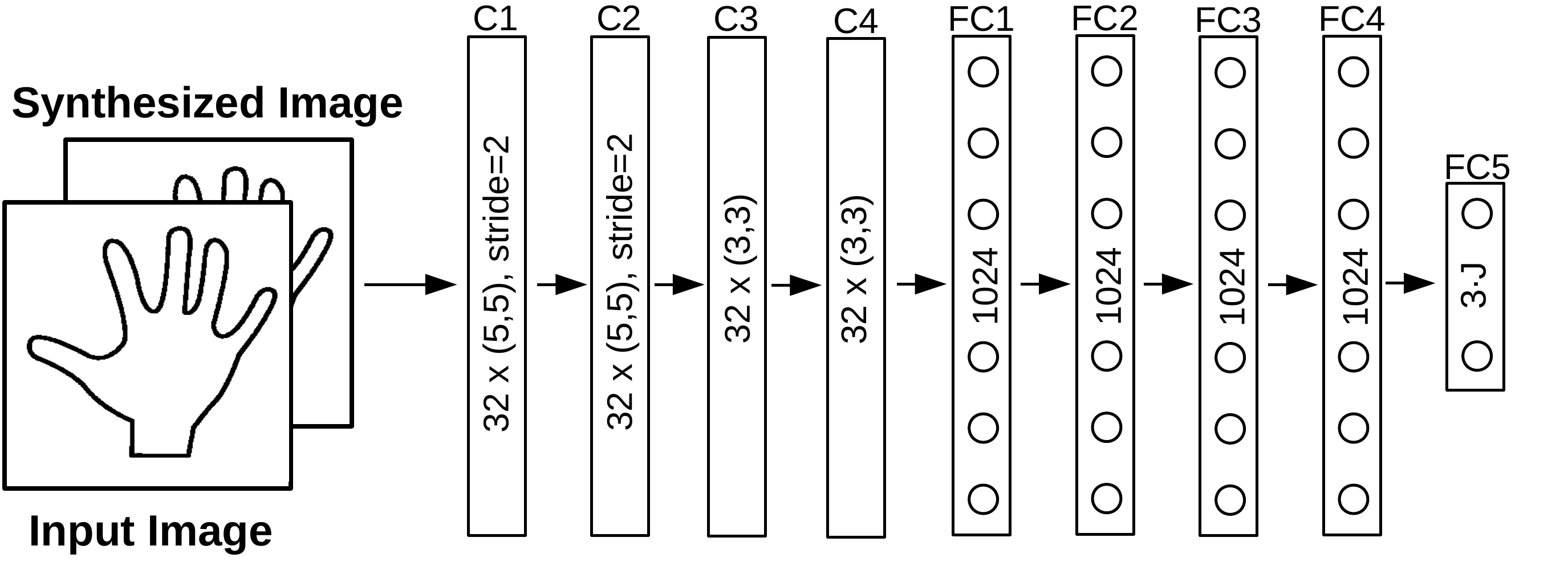}
\end{center}
   \caption{Network architecture of the updater CNN. The network contains several convolutional  layers that
     use a  filter stride  to reduce the  size of the  feature maps.   The final
     feature maps are fed  into a fully connected network.  All
     layers have rectified-linear units, except  the last layer which has linear
     units.  The pose update is used to  refine the initial pose and the refined
     pose is again fed into the  synthesizer CNN to iterate the whole procedure.  As
     in Fig.~\ref{fig:architecture_synth},  \textsf{C} denotes a  convolutional layer,
     and \textsf{FC} a fully connected layer.}
\label{fig:architecture_refine_comp}
\label{fig:updater}
\end{figure}

The  updater CNN  $\updater(.,.)$  takes  two depth  images  as input.   As
already stated  in Eq.~\eqref{eq:updater}, at  run-time, the first image  is the
input depth image, the second image is the image returned by the synthesizer CNN for
the  current pose  estimate. Its  output  is an  update that  improves the  pose
estimate. The architecture is shown in Fig.~\ref{fig:updater}. The input to the network is the observed image stacked channel-wise with the  image from  the synthesizer CNN.  We do not  use max-pooling here,  but a
filter stride~\cite{Jain2014,Liu2015} to reduce the resolution of the feature maps. We experienced inferior
accuracy  with  max-pooling compared  to that with stride,  probably because  max-pooling
introduces spatial  invariance~\cite{Scherer2010} that  is not desired  for this
task.

\begin{figure}
\begin{center}
\includegraphics[width=0.6\linewidth]{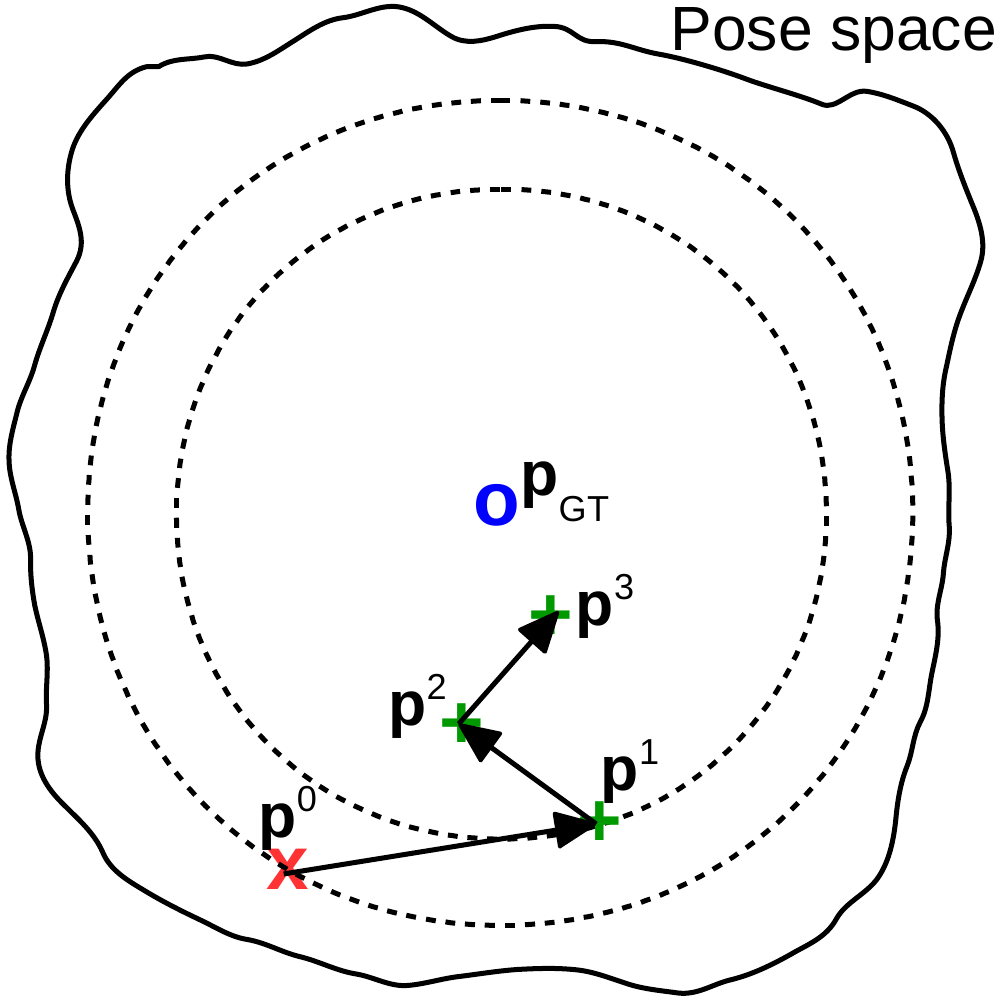}
\end{center}
   \caption{Our   iterative  pose   optimization   in  high-dimensional   space,
     schematized    here   in    2D.    We    start   at    an   initial    pose
     (\textcolor{red}{$\bm{\times}$}) and  want to converge to  the ground truth
     pose  (\textcolor{blue}{$\bm{\circ}$}),  that maximizes  image  similarity.
     Our updater CNN generates updates for each pose (\textcolor{green}{$\bm{+}$})
     that  bring us  closer. The  updates are  predicted from  the synthesized image of the current  pose
     estimate and the observed depth image.}
\label{fig:iterative}
\end{figure}

Ideally, the output of the updater CNN should  bring the pose  estimate to the  correct pose  in a
single step.  This is a very  difficult problem, though, and we  could not get the network to reduce the initial training error within a reasonable timeframe. 
However, our only requirement from the updater CNN is for it to
output an update  which brings us closer  to the ground truth as shown in Fig.~\ref{fig:iterative}.   We iterate
this  update procedure  to get  closer  step-by-step.  Thus,  the update  should
follow the inequality
\begin{equation}\label{eq:train_refine2}
	\lVert \pose + \updater(\calD, \synth(\pose)) -
        \pose_\text{GT} \rVert < \lambda \lVert \pose
        -\pose_\text{GT} \rVert \;\; ,
\end{equation}
where $\pose_\text{GT}$ is the ground truth pose for image $\calD$, and ${\lambda
\in [0,1]}$ is  a multiplicative factor that specifies a  minimal improvement. We
use  $\lambda =  0.6$  in  our experiments.

We  optimize the  parameters $\Omega$ of the updater CNN by minimizing the following cost function
\begin{equation}\label{eq:train_refine}
	\mathcal{L} = \sum_{(\calD,\pose)\in\calT} \sum_{\pose'\in \calT_{\calD}} \max(0,\lVert \pose'' - \pose \rVert
  - \lambda \lVert \pose' -\pose \rVert) \;\; , 
\end{equation}
where   $\pose''   =   \pose'  +   \updater_\Omega(\calD,\synth(\pose'))$,   and
$\calT_\calD$ is a set of poses.   The introduction of the synthesizer CNN allows us
to   virtually  augment   the  training   data  and   add  arbitrary   poses  to
$\calT_{\calD}$, which the updater CNN is then trained to correct.

The set $\calT_{\calD}$ contains the ground truth $\pose$, for which the updater CNN
should output a zero update.  We  further add as many meaningful deviations from
that ground truth  as possible, which the updater CNN might perceive during testing and be
asked  to correct.   We start  by  including the  output pose  of the  predictor CNN
$\pred(\calD)$, which   is used during testing as  initialization of  the update
loop.  Additionally, we  add copies  with small  Gaussian
noise for all  poses.  This  creates convergence basins around  the ground truth, in  which the
predicted updates point towards the ground  truth, as we show in the evaluation,
and helps to explore the pose space.

After every 2 epochs, we augment the set by applying the current updater CNN to
the poses in $\calT_\calD$, that is, we permanently add the set
\begin{equation}
\label{eq:train_update_poses}
\{\pose_2 \;|\; \exists \pose \in \calT_\calD  \text{ s.t. } \pose_2 = \pose + \updater(\calD, \synth(\pose)) \} \; 
\end{equation}
to $\calT_\calD$. This extends the training set with poses made of the predicted updates from the current training set $\calT_\calD$.
This forces the updater CNN to learn to further improve on its own outputs.

In addition, we sample from the current distribution of errors across all the samples,
and add these errors to the poses, thus explicitly focusing the training on common deviations.
This is  different from the  Gaussian noise  and helps to  predict correct
updates for larger initialization errors.

\subsection{Learning all Functions Jointly}
So far, we have considered optimizing all functions separately, \ie, first
training the synthesizer CNN and the predictor CNN, and then using them to train
the updater CNN. However, it is theoretically possible to train the three
networks together. One iteration can be expressed in terms of the current pose $\pose$ by introducing the following $\iter(.)$ function:
\begin{equation}
\label{eq:train_jointly1}
	\iter(\pose) = \pose+\updater(\synth(\pose),\calD) \; \; .
\end{equation}
The pose estimate $\widehat\pose^{(n)}$ after $n$ iterations can be written as:
\begin{equation}
\label{eq:train_jointly2}
	\widehat\pose^{(n)} = \iter(\hdots \iter(\pred(\calD))\hdots) \; \; .
\end{equation}
$\pred(.)$, $\synth(.)$, and $\updater(.)$ can now be trained by minimizing
\begin{equation}
	\{\widehat{\Phi}, \widehat{\Theta}, \widehat{\Omega}\} = \argmin_{\{\Phi,\Theta,\Omega\}} \sum_{(\calD, \pose) \in \calT} \lVert \widehat\pose^{(n)} - \pose\rVert^2 \; \; .
\label{eq:jointly}
\end{equation}

The function $\iter(.)$ can be seen as a Recurrent Neural Network (RNN) that
depends on the input depth image $\calD$. In comparison to RNNs, our method
makes training simpler, intermediary steps easier to understand, and the design of
the networks' architectures easier.

We tried to optimize Eq.~\eqref{eq:jointly} starting from a random network initialization, but the optimization did not converge to a satisfying solution. Also, when using the pretrained synthesizer CNN and predictor CNN as initialization, the optimization converges, but this leads to similar accuracy, indicating that end-to-end optimization is possible, but not useful here. Moreover, the intermediate results,~\ie, the synthesized depth images, are less interpretable as the hand is barely recognizable. Splitting the optimization problem as we do makes therefore the optimization easier, at no loss of accuracy.

\section{Joint Hand-Object Pose Estimation}
\label{sec:main_ho}
\begin{figure*}
\includegraphics[width=0.95\linewidth]{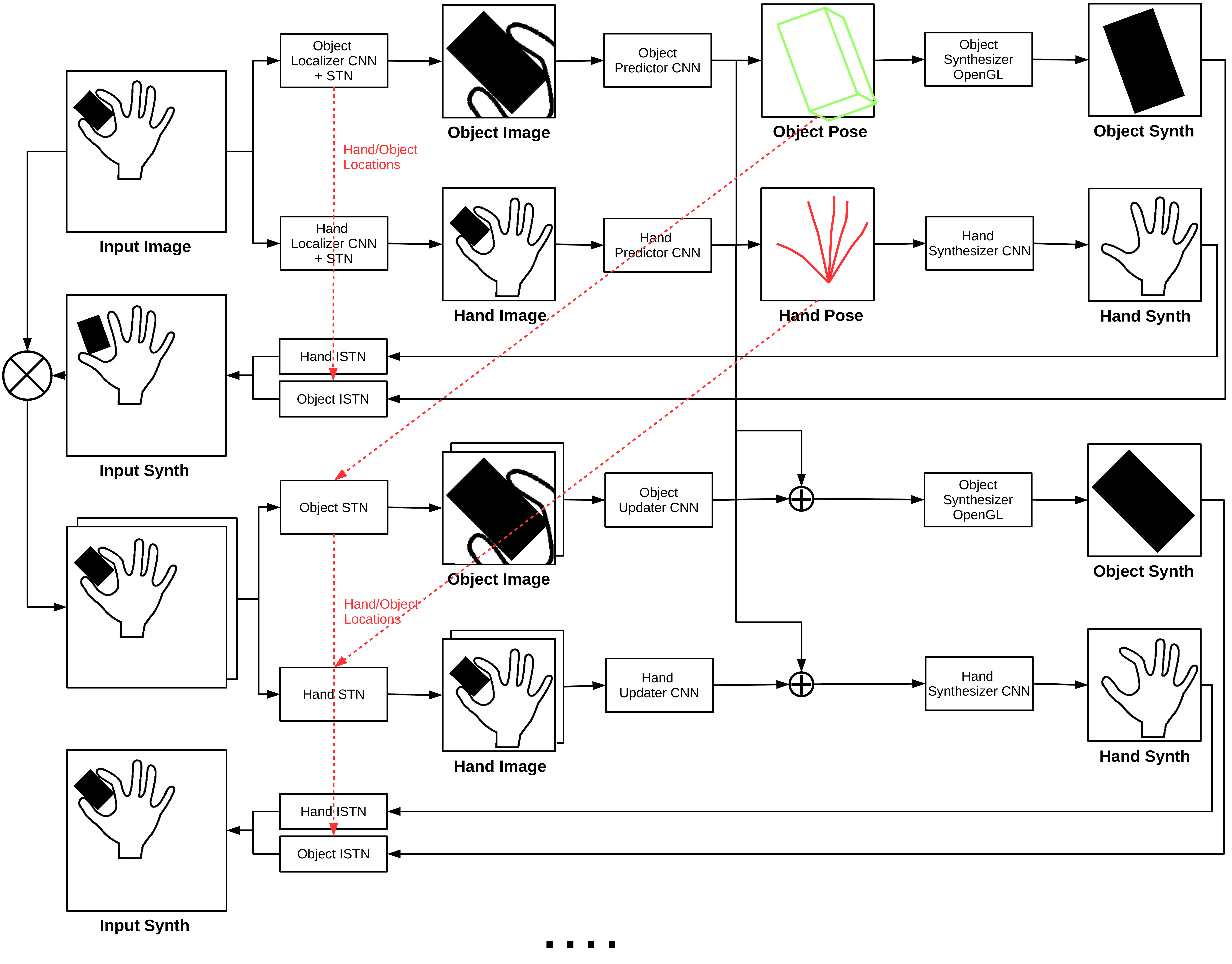}
\caption{Overview of our joint hand-object pose estimation method. We show the predictors for the initial poses with one iteration of the updaters. The input to our method is a crop from the depth camera frame that contains the hand with the object estimated from the center-of-mass in the depth camera frame. From this input, we localize the hand and the object separately, by predicting the 2D locations and depth using the localizer CNNs. We apply a Spatial Transformer Network~(STN) to crop a region of interest around the predicted location. On this centered crop, we predict an initial pose using the predictor CNNs, which are then used to synthesize depth images of the hand and the object using the synthesizer CNNs. Using an Inverse STN~(ISTN), we paste back the different inputs,~\ie, the synthesized object and the synthesized hand, onto the input image. These images are stacked together ($\otimes$) and serve as input to the updater CNNs that predict one update for the pose of the hand and one for the object. The updates are added to the poses and the procedure is iterated several times.}
\label{fig:overview_ho}
\end{figure*}

We now aim at estimating both the pose $\pose^H$ of a hand and the pose $\pose^O$ of an object simultaneously while the hand is manipulating the object. We start the pose prediction by first estimating the locations of the hand and the object within the input image. Using the localization, we predict an initial estimate of the pose for the hand and the object separately. In practice, these poses are not very accurate, since hand and object can severely occlude each other. Therefore, we introduce feedback by first synthesizing depth images of the hand and the object, and merge them together. An overview of our method with the different operations and networks is shown in Fig.~\ref{fig:overview_ho}.

We can then predict an update that aims at correcting the object pose and the hand pose.
Each step is performed by a CNN. As in the previous section, we denote them localizer CNN $\loc(.)$, predictor CNN $\pred(.)$, synthesizer CNN $\synth(.)$, and updater CNN $\updater(., .)$. However, this time we use one network for the hand and one for the object and use suffixes $^H$ and $^O$, respectively, to distinguish them. In the following we detail how the individual networks are trained.

Since we now aim at estimating the hand pose and the object pose simultaneously, we require for the training set the hand pose and the object pose for each depth frame $\calT = \{(\calD_i,  \pose^O_i, \pose^H_i)\}^N_{i=1}$. Acquiring these annotations for real data can be very cumbersome, and therefore we only use synthetically generated training data, as detailed in the next section. 

\subsection{Training Data Generation}
\label{sec:training_data}

\begin{figure}
{\tabcolsep=0pt
\begin{tabular}{cc}
\includegraphics[width=0.45\linewidth,clip,trim={3cm 3cm 3cm 2cm}]{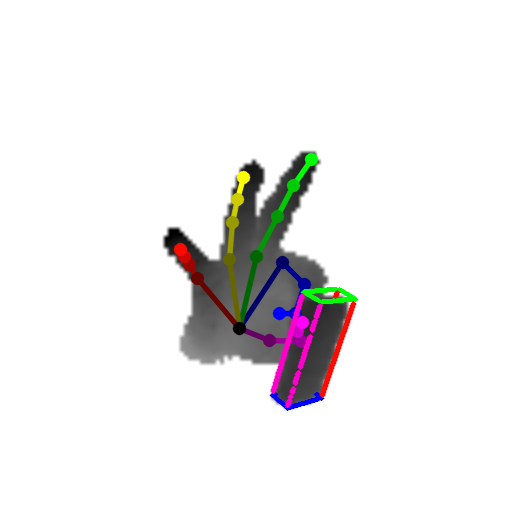} &
\includegraphics[width=0.45\linewidth,clip,trim={3cm 4cm 3cm 7cm}]{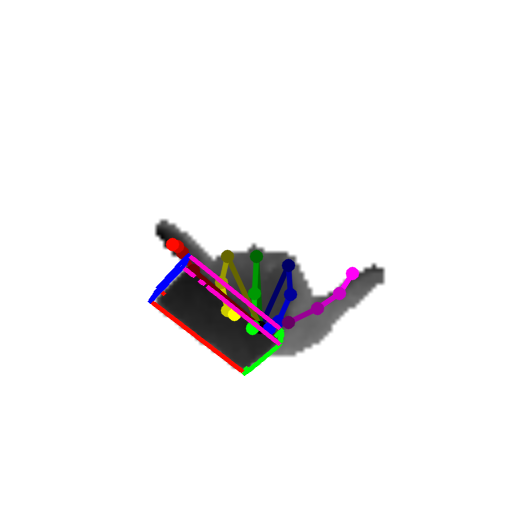} \\
\includegraphics[width=0.22\linewidth,clip,trim={4cm 2cm 7cm 4cm}]{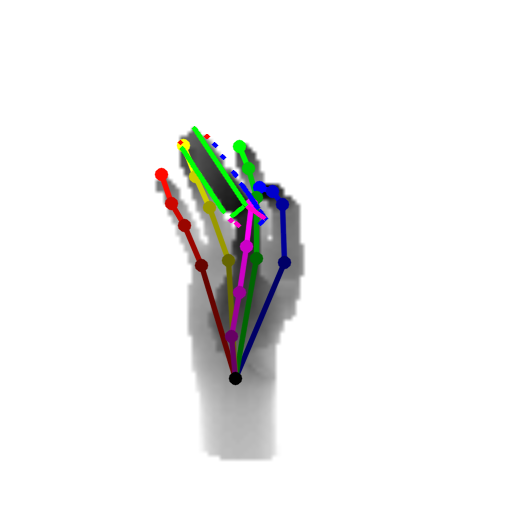} &
\includegraphics[width=0.22\linewidth,clip,trim={4cm 2cm 8cm 4cm}]{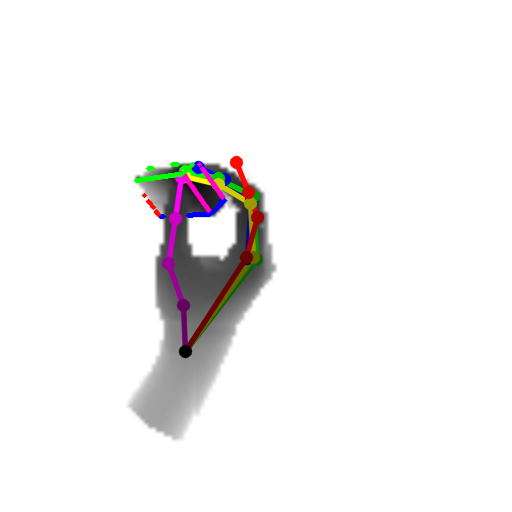} \\
\end{tabular}
}
\caption{Training samples used for joint hand-object pose estimation. \textbf{Top row:} Semi-synthetic training data comprising real depth images of hands with rendered objects. The interaction between hand and object cannot be \textit{accurately} modeled easily, but the solution space of possible object locations and possible object poses can be significantly reduced, which enables training our proposed method. \textbf{Bottom row:} Additional synthetic depth data is used to accurately model the hand-object interaction, but it does not capture the sensor characteristics.}
\label{fig:training_samples}
\end{figure}

Capturing real training data of a hand manipulating an object can be very cumbersome, since in our case it requires 3D hand pose and 3D object pose annotations for each frame. This is hindered by severe occlusions, or not always possible at all. Our approach leverages annotations from datasets of hands in isolation, which are much simpler to capture in practice~\cite{Sun2015,Qian2014,Tang2014}, and data from 3D object models, which are available at large scales~\cite{Chang2015}. We use an OpenGL-based rendering of the 3D object model and fuse the rendering with the frames from a 3D hand dataset. Since we use depth images, we can simply take the minimum of both depth images for each pixel. We render the object on top of the hand, placing the object near the fingers, and apply a simple collision detection, such that the object is not placed ``within'' the hand point cloud. The object poses,~\ie, rotations and translations, are sampled randomly, constrained by the collision detection. In order to account for sensor noise, we add small Gaussian noise to the rendering. In Fig.~\ref{fig:training_samples} we show some samples of synthesized training data. 

Besides this semi-synthetic training data, we also use completely synthetic training data of hands and objects. We therefore use a marker-based motion capture system~\cite{Han2018} to capture the hand and object pose while the hand is manipulating different objects. Since the image data is corrupted with the markers, we render a synthetic hand model and the CAD object model given the captured poses, and use this rendering as additional training data. 
This dataset comprises five users performing gestures with five different hand-held objects. In total, 73k frames are synthetically rendered from the captured poses from different viewpoints and used for training.

\subsection{Learning the Hand and Object Localizers}
\label{sec:ho_loc}
During experiments we have found that accurate localization of the hand and the object is crucial for accurate pose estimation. Since we jointly consider a hand and an object, we use the combined center-of-mass to perform a rough localization and extract a larger cube around the center-of-mass that contains the hand and the object. Then, we use two CNNs to perform the localization of the hand and the object separately. We train two networks, $\loc^H(.)$ for the hand, and $\loc^O(.)$ for the object, both taking the same input image cropped from the center-of-mass location. The network architectures are the same as the one used in Section~\ref{sec:loc} and shown in Fig.~\ref{fig:architecture_loc}. The networks are trained to predict the 2D location and depth of the MCP joint of the hand $\mathbf{l}^H$ and the object centroid $\mathbf{l}^O$, respectively, relative to the center-of-mass.
Formally, we optimize the cost function
\begin{equation}
\widehat{\Gamma} = \argmin_\Gamma \sum_{(\calD, \mathbf{l}^H) \in \calT} \lVert \loc_\Gamma^H(\calD) - \mathbf{l}^H\rVert^2
\end{equation}
for the hand localizer, and
\begin{equation}
\widehat{\chi} = \argmin_\chi \sum_{(\calD, \mathbf{l}^O) \in \calT} \lVert \loc_\chi^O(\calD) - \mathbf{l}^O\rVert^2
\end{equation}
for the object localizer, where $\Gamma$ and $\chi$ are the parameters of the localizer CNNs.

\subsection{Spatial Transformer and Inverse Spatial Transformer Networks}
\label{sec:stn}
Once we are given the location of the hand and the object, we use them to crop a region of interest around the hand and the object, respectively, using a Spatial Transformer Network~(STN). 
Note that this makes the crop differentiable and thus trainable end-to-end, and allows faster inference on the GPU since we can run a single large network made from the individual CNNs.
The STN proposed in~\cite{Jaderberg2015} was parametrized to work with 2D images. In this work, we apply the STN on 2.5D depth images. We estimate the spatial transformation $A_\theta$ from the 3D location, calculated from the predicted 2D location and depth. Using the intrinsic camera calibration we denote the projection from 3D to 2D as $\proj(\cdot)$, the 3D bounding box as $c$, and the 3D location as $\mathbf{t}$. We can estimate the 2D bounding box projection as $(x^\pm, y^\pm) = \proj(\mathbf{t}\pm c)$ and define $x^\Delta = (x^+ - x^-)$ and $y^\Delta = (y^+ - y^-)$. This leads to the parametrization of the spatial transformation $A_\theta$ as:
\begin{equation}
A_\theta = 
\begin{pmatrix}
x^\Delta/2 & 0 & x^- + x^\Delta/2 \\
0 & y^\Delta/2 & y^- + y^\Delta/2 \\
0 & 0 & 1
\end{pmatrix} \; \; ,
\end{equation}
for the parametrization of the sampling grid locations as in~\cite{Jaderberg2015}.
The source sampling grid locations $(x^s, y^s)$ are obtained by transforming the regularly sampled target grid locations $(x^t, y^t)$ as:
\begin{equation}
(x^s, y^s)^T = A_\theta \cdot (x^t, y^t, 1)^T \; \; .
\end{equation}
We then use bilinear sampling to interpolate the depth values of input depth image $\calD$ of size $W\times H$:
\begin{multline}
\STN(\mathbf{l}, \calD) = \\
 \sum^H_h \sum^W_w \calD[h,w] \max(0, 1-|x^s-w|)\max(0, 1-|y^s-h|) \; \; .
\end{multline}

Since we apply a detection and crop using the STN, we require the inverse operation for putting the synthesized object back into the original image frame before applying the updater CNN. Hence, we apply an Inverse STN~(ISTN)~\cite{Lin2017}. It operates the same way as the STN, but uses the inverse of the spatial transformation $A_\theta$.

\subsection{Learning the Hand and Object Pose Predictors}
Once we are given an accurate location of the hand and the object, we crop a patch around the hand and the object using the STN and use the crop to estimate the pose. We denote $c^H$ the crop of the hand, and $c^O$ the crop of the object. We denote this operation as $c^O = \STN(\loc^O(\calD), \calD)$.
For the pose prediction, we require appropriate parametrization of the hand and object poses. For the object pose $\pose^O$, we use the eight corners of its 3D bounding box, which was shown to perform better than rotation and translation~\cite{Xiang2018}. For the hand pose, we use the 3D joint locations with a prior, as in Section~\ref{sec:pred}, on the 3D hand poses obtained from the data described in Section~\ref{sec:training_data}. A prior on the 3D hand pose is very effective in case of occlusions, since it constraints the predicted hand pose to valid poses. For our experiments with hands manipulating objects we use a similar architecture to the network shown in Fig.~\ref{fig:architecture_loc} for the sake of speed. However, for the hand predictor CNN $\pred^H(.)$ we use $3\cdot J$ neurons for the output layer and for the object predictor CNN $\pred^O(.)$ we use $3\cdot 8$ neurons. \\
Again, we optimize two cost functions to train the predictor CNNs. For the hand pose predictor we optimize
\begin{equation}
\widehat{\eta} = \argmin_\eta \sum_{(\calD, \pose^H) \in \calT} \lVert\mathbf{P}^{-1}(\pred_\eta^H(c^H)) - \pose^H \rVert^2 \; \; ,
\end{equation}
where $\mathbf{P}$ is the linear prior on the 3D hand poses.
For the object pose predictor we optimize
\begin{equation}
\widehat{\Pi} = \argmin_\Pi \sum_{(\calD, \pose^O)\in \calT} \lVert \pred_\Pi^O(c^O) - \pi(\pose^O) \rVert^2 \; \; ,
\end{equation}
where $\pi(\cdot)$ returns the 3D bounding box coordinates given the object pose.
$\eta$ and $\Pi$ denote the parameters of the predictor CNNs.
We use orthogonal Procrustes analysis~\cite{Kabsch1976} to obtain the 3D object pose from the output of $\pred^O(.)$.

\subsection{Learning the Hand and Object Updaters}
Since we now have initial estimates of the pose for the hand and the object, we can use them to start a feedback loop that considers the interaction of hand and object. Therefore, we extend the input data of the updater CNN with the synthesized image of the hand and the object given the initially predicted poses. This requires training a synthesizer CNN, or using a software renderer. Since we have the 3D model of the object readily available, we implement the object synthesizer $\synth^O(.)$ with an OpenGL-based renderer. During our experiments, we also evaluated training the updater CNN with the object synthesizer CNN, and this resulted in similar results compared to the OpenGL-based rendering. The hand synthesizer CNN $\synth^H(.)$ is trained in the same way as described in Section~\ref{sec:synth}. Also, the network architecture of the updater CNN is the same as shown in Fig.~\ref{fig:architecture_refine_comp} in the previous section.

For training the updater CNN, the synthesized images are merged into a single depth image:
\begin{equation}
\begin{split}
S = \min( & \ISTN(\loc^H(\calD), \synth(\pose^H)), \\
 & \ISTN(\loc^O(\calD), \synth(\pose^O))) \otimes \calD \; \; , 
\end{split}
\end{equation}
where $\min(A, B)$ denotes the pixel-wise minimum between two depth images $A$ and $B$, and $\otimes$ denotes a channel-wise stacking. We then train the updater CNN for the hand $\updater^H(., .)$ and the object $\updater^O(., .)$ by minimizing the following cost function:
\begin{equation}
\label{eq:refine_ho}
\sum_{(\calD, \pose^H, \pose^O)\in \calT} \sum_{\pose'^{O} \in \calT_O} \sum_{\pose'^{H} \in \calT_H} \max(0, \lVert \pose'' - \pose \rVert - \lambda \lVert \pose'^{H,O} - \pose\rVert) \; \; ,
\end{equation}
where $\pose'' = \pose'^{H,O} + \updater^{H,O}(c^{H,O}, S)$, and $\calT_H$ and $\calT_O$ are sets of poses for the hand and the object, respectively. When optimizing Eq.~\eqref{eq:refine_ho} for the hand updater $\updater^H(.,.)$, the variables with suffix $^H$ are used, and when optimizing for the object updater $\updater^O(.,.)$ the variables with suffix $^O$ are used.\\
When training the updater CNN $\updater^H(.,.)$ for the hand, $\calT_H$ is initialized and updated during training as described in Section~\ref{sec:feedback}. $\calT_O$ is made from random poses around the ground truth object pose, and poses from the predictor $\pred^O(.)$ applied to training images. \\
When training the updater CNN $\updater^O(.,.)$ for the object, $\calT_O$ is defined as in Section~\ref{sec:feedback}. $\calT_H$ is made from random poses around the ground truth together with predictions from the training data. 

For inference, we iterate the updater CNNs several times: We first obtain the
initial estimates $\pose^{(0)}$ for hand $H$ and object $O$, by running the
predictor CNNs on the cropped locations from the localizer CNNs. Then we predict
the updates using the merged image $S^{(i)}$ from the previous iteration.
Alg.~\ref{fig:algorithm} gives a formal expression of this algorithm.

\begin{algorithm}
\caption{Feedback Loop}
\begin{algorithmic}[1]
\State{$\widehat\pose^{(0),H} \leftarrow \pred^H(\STN(\loc^H(\calD), \calD)$}
\State{$\widehat\pose^{(0),O} \leftarrow \pred^O(\STN(\loc^O(\calD), \calD)$}
    \For{$i \gets 0$ to $N-1$}
    \State{$\begin{aligned}
S^{(i)} \leftarrow \min( &\ISTN(\loc^H(\calD), \synth(\widehat\pose^{(i),H})), \\
& \ISTN(\loc^O(\calD), \synth(\widehat\pose^{(i),O}))) \otimes \calD
\end{aligned}$}
\State{$\widehat\pose^{(i+1),H} \leftarrow  \widehat\pose^{(i),H} + \updater^H(\STN(\pose^{(i),H}, \calD), S^{(i)})$}
\State{$\widehat\pose^{(i+1),O} \leftarrow  \widehat\pose^{(i),O} + \updater^O(\STN(\pose^{(i),O}, \calD), S^{(i)})$}
   \EndFor
\end{algorithmic}
\label{fig:algorithm}
\end{algorithm}

\section{Hand Pose Evaluation}
\label{sec:eval}
In  this  section  we  evaluate  our  proposed  method  on  the  NYU  Hand  Pose
Dataset~\cite{Tompson2014},  a challenging  real-world benchmark  for hand  pose
estimation. First, we describe  how we train the  networks.  Then, we introduce the  benchmark  dataset.  Furthermore,  we evaluate  our method qualitatively and quantitatively.

\subsection{Training}

We optimize the network parameters using gradient descent, specifically using the ADAM method~\cite{Kingma2015} with default hyper-parameters.
  The batch  size is $64$.   The learning
rate  decays over  the  epochs and  starts  with $0.001$. The  networks are  trained for
$100$ epochs. We augment the training data online during training by random scales, random crops, and random rotation~\cite{Oberweger2017}.

We extract  a fixed-size metric cube  from the depth image  around the
hand location. The depth values within the cube are resized to a $128\times128$
patch and normalized to $[-1,1]$. The  depth values are clipped to the
cube sides  front  and  rear. Points  for  which  the  depth  is
undefined---which  may  happen  with   structured  light  sensors  for
example---are assigned  to the  rear side.  This preprocessing  step
was also done in~\cite{Tang2014} and provides invariance to different hand-to-camera distances.

\subsection{Benchmark Dataset}
We evaluated our  method on the NYU Hand  Pose Dataset~\cite{Tompson2014}.  This
dataset is  publicly available and it  is backed up by a huge quantity of annotated  samples together with very accurate annotations.  It also shows a high variability  of poses, however, which can
make  pose estimation  challenging.   While the  ground  truth contains  $J=36$
annotated      joints,      we      follow     the      evaluation      protocol
of~\cite{Oberweger2015,Tompson2014} and use the same subset of $J=14$ joints.

The training set contains samples of one  person, while the test set has samples
from  two persons.   The dataset  was captured  using a  structured light  RGB-D
sensor, the Primesense Carmine 1.09, and  contains over 72k training and 8k test
frames. We  used only the depth  images for  our experiments. They  exhibit  typical artifacts    of structured light  sensors: The outlines  are noisy  and there are  missing depth
values along occluding boundaries.

\subsection{Comparison with Baseline}

We show the benefit of using our proposed feedback loop to increase the accuracy of the 3D joint localization.
For this, we   compare    our   method   to   very    recent   state-of-the-art   methods:
\textit{DeepPrior++}~\cite{Oberweger2017} integrates a prior  on the 3D hand poses
into a  Deep Network;  \textit{REN}~\cite{Guo2017} relies on  an ensemble  of Deep
Networks,    each   operating    on    a   region    of    the   input    image;
\textit{Lie-X}~\cite{Xu2016} uses  a sophisticated tracking  algorithm constrained
to  the  Lie  group;  \textit{Crossing Nets}~\cite{Wan2017}  uses  an  adversarial
training architecture; Neverova~\etal~\cite{Neverova2017}  proposed a
semi-supervised approach  that incorporates a semantic segmentation of  the hand;
\textit{DeepModel}~\cite{Zhou2016}  integrates  a  3D  hand  model  into  a  Deep  Network;
\textit{DISCO}~\cite{Bouchacourt2016}  learns the  posterior  distribution  of hand  poses;
\textit{Hand3D}~\cite{Deng2017} uses a volumetric CNN to process a
point cloud, similar to \textit{3DCNN}~\cite{Ge2018}.

We show quantitative comparisons in Table~\ref{tab:nyu_quantitative}, which compares the different methods we consider using the average Euclidean distance between ground truth and predicted joint 3D locations, which is a \textit{de facto} standard for this problem. For our feedback loop, we use DeepPrior++~\cite{Oberweger2017} for initialization as described in Section~\ref{sec:pred}, which performs already very accurately. Still, our feedback loop can reduce the error from 12.2~mm to 10.8~mm.

\begin{table}
\caption{Quantitative evaluation on the NYU dataset~\cite{Tompson2014}. We compare the average Euclidean 3D error of the predicted poses with state-of-the-art methods on the NYU dataset.}
\label{tab:nyu_quantitative}
\centering
\begin{tabular}{@{}lcc@{}}
\toprule
Method & & Average 3D error \\
\midrule
Neverova~\etal~\cite{Neverova2017} && 14.9mm \\
Crossing Nets~\cite{Wan2017} && 15.5mm \\
Lie-X~\cite{Xu2016} && 14.5mm \\
REN~\cite{Guo2017} && 13.4mm \\ 
DeepPrior++~\cite{Oberweger2017} && 12.3mm\\
Feedback~\cite{Oberweger2015a}  && 16.2mm \\
Hand3D~\cite{Deng2017} && 17.6mm \\
DISCO~\cite{Bouchacourt2016} && 20.7mm \\
DeepModel~\cite{Zhou2016} && 16.9mm \\
Pose-REN~\cite{Chen2017} && 11.8mm \\
3DCNN~\cite{Ge2018} && 10.6mm \\
\midrule
Ours && 10.8mm \\
\bottomrule
\end{tabular}
\end{table}

\subsection{Image-Based Hand Pose Optimization}
\label{sec:naive}

We  mentioned   in  Section~\ref{sec:overview}   that the attempt may be made
to estimate the pose by directly optimizing the squared loss between
the input image and the synthetic one as given in Eq.~\eqref{eq:naive} and we
argued that this does not in fact work well. We now demonstrate this empirically.

We  used  the  powerful  L-BFGS-B  algorithm~\cite{Byrd1995},  which  is  a  box
constrained optimizer, to solve Eq.~\eqref{eq:naive}.  We set the constraints on
the pose in such a manner that each joint coordinate stays inside the hand cube.

The minimizer  of Eq.~\eqref{eq:naive},  however, does not correspond  to a better
pose in general, as  shown in Fig.~\ref{fig:gensamplesBFGS}. Although  the generated image
looks very similar to the input image, the pose does not improve, moreover it even often becomes worse. Several reasons can account for this. The depth  input image
typically exhibits noise along the contours, as  in the example 
of  Fig.~\ref{fig:gensamplesBFGS}. After  several  iterations  of L-BFGS-B,  the
optimization may start corrupting the pose estimate with the result that the synthesizer generates artifacts that fit the noise. We quantitatively evaluated the image-based optimization and it gives an average Euclidean error of 32.3~mm on the NYU dataset, which is actually worse than the initial pose of the predictor CNN, which achieves 12.2~mm error.

Furthermore  the  optimization is  prone  to  local minima  due  to  a noisy  error
surface~\cite{Qian2014}.     However,    we    also   tried    Particle    Swarm
Optimization~\cite{Oikonomidis2011a,Qian2014,Sharp2015}   a  genetic   algorithm
popular for hand pose optimization, and  obtained similar results. This tends to
confirm  that  the  bad  performance   comes  from  the  objective  function  of
Eq.~\eqref{eq:naive} rather than the optimization algorithm.

\begin{figure}
\begin{center}
    \includegraphics[width=0.75\linewidth]{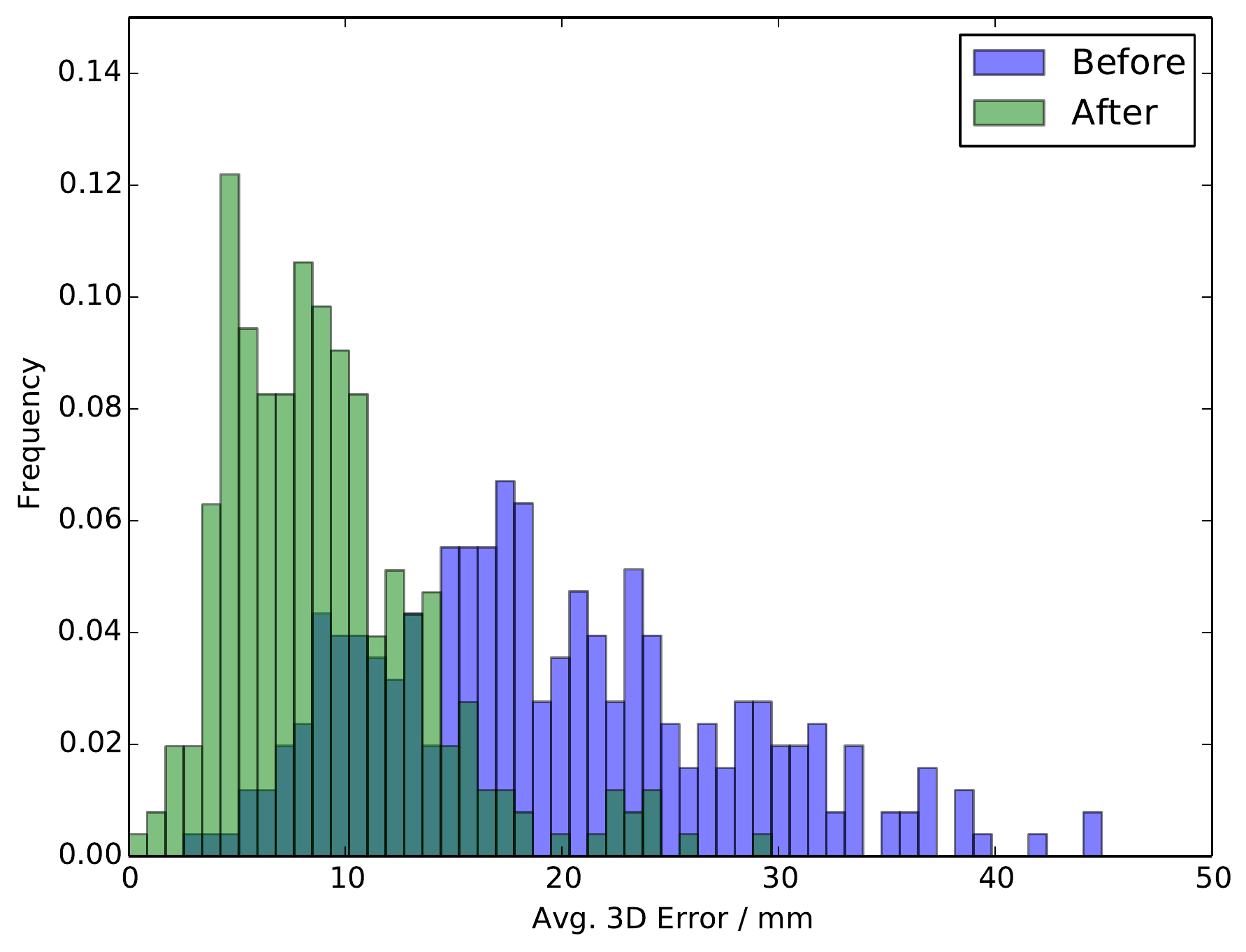} 
\end{center}
   \caption{Histogram of errors before and after applying the predicted updates. We initialize noisy joint locations around the ground truth location and calculate the average 3D error before and after applying the update. The different updates minimize the error and thus bring us closer to the ground truth pose.}
\label{fig:results_hist}
\end{figure}

We further show the histogram of average 3D errors before and after applying the predicted updates for different initializations around the ground truth location in Fig.~\ref{fig:results_hist}. As evident from the distribution of errors, the updater improves all initializations.

\bgroup
\setlength{\tabcolsep}{1pt}
\begin{figure}[t]
\begin{center}
\begin{tabular}{@{}ccccc@{}}
Init & Init & Iter 1 & Iter 2 & Final \\
\multirow{2}{*}[2em]{
\includegraphics[width=0.19\linewidth, trim={2cm 2cm 0cm 0cm},clip]{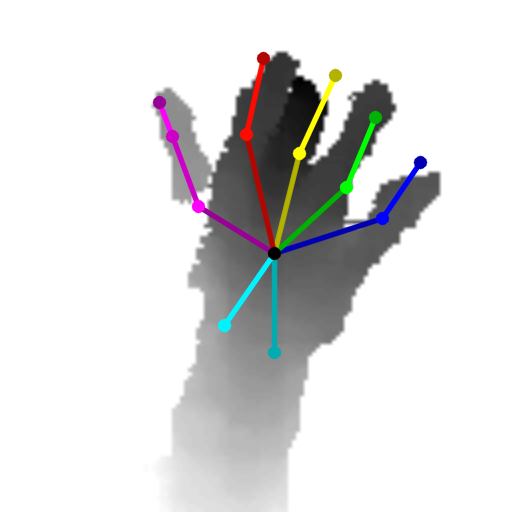}} &
\includegraphics[width=0.19\linewidth, trim={0.5cm 0.5cm 0cm 0cm},clip]{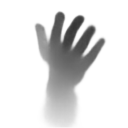} &
\includegraphics[width=0.19\linewidth, trim={0.5cm 0.5cm 0cm 0cm},clip]{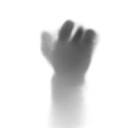} &
\includegraphics[width=0.19\linewidth, trim={0.5cm 0.5cm 0cm 0cm},clip]{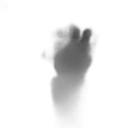} &
\includegraphics[width=0.18\linewidth, trim={0.5cm 0.5cm 0cm 0cm},clip]{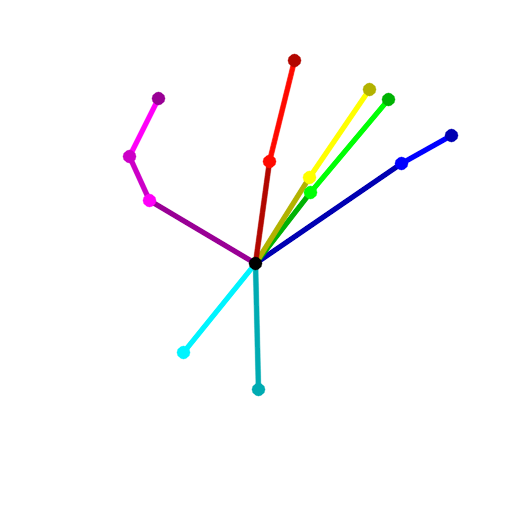} \\

 &
\includegraphics[width=0.19\linewidth, trim={0.5cm 0.5cm 0cm 0cm},clip]{GENREF_geninit_129_iter0} &
\includegraphics[width=0.19\linewidth, trim={0.5cm 0.5cm 0cm 0cm},clip]{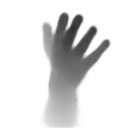} &
\includegraphics[width=0.19\linewidth, trim={0.5cm 0.5cm 0cm 0cm},clip]{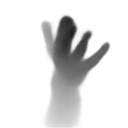} &
\includegraphics[width=0.17\linewidth, trim={0.5cm 0.5cm 0cm 0cm},clip]{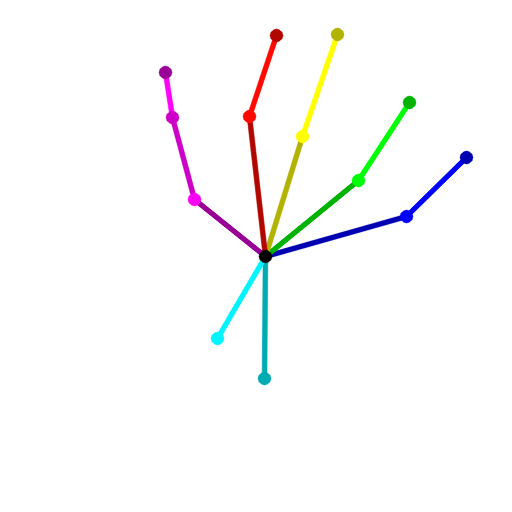} 
\end{tabular}
\end{center}
   \caption{Comparison with image-based  pose optimization. \textbf{(Top)} results for image-based optimization, and \textbf{(bottom)} for our proposed method.    From left to right:  input depth image with initial pose, synthesized image for initial pose, after first, second iteration, and final pose. Minimizing the difference between the synthesized and the input  image does not induce better  poses due to sensor noise and local minima. Thanks to the updater, our method can fit a good estimate.}
\label{fig:gensamplesBFGS}
\end{figure}
\egroup

\subsection{Qualitative results}

\bgroup
\setlength{\tabcolsep}{1pt}
\begin{figure*}[t]
\begin{center}
\newlength{\sampimgw}
\setlength\sampimgw{0.12\linewidth}
\begin{tabular}{@{}cc|cc|cc||cc@{}}
\includegraphics[width=\sampimgw,trim={3cm 1cm 3cm 1cm},clip]{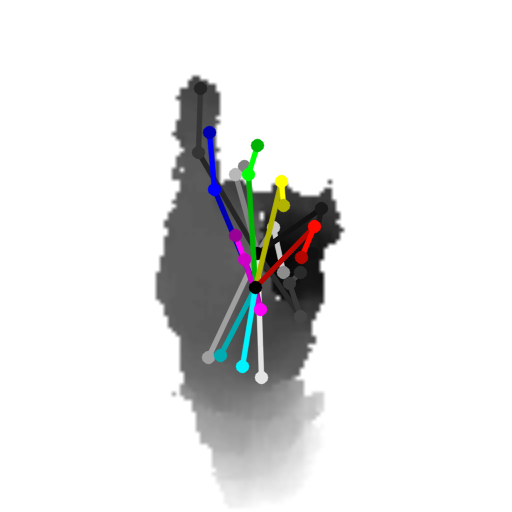} &
\includegraphics[width=\sampimgw,trim={3cm 1cm 3cm 1cm},clip]{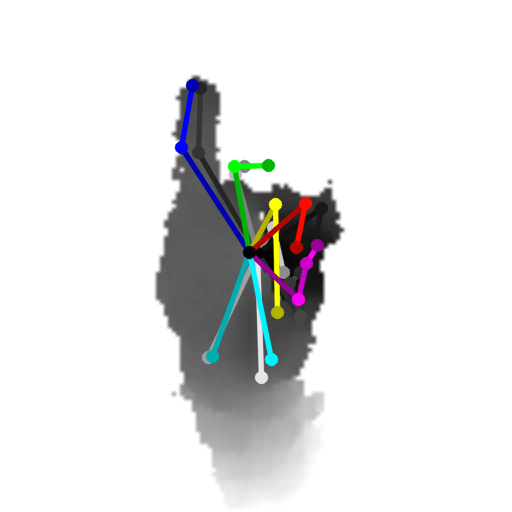} &
\includegraphics[width=\sampimgw,trim={3cm 2cm 2cm 1cm},clip]{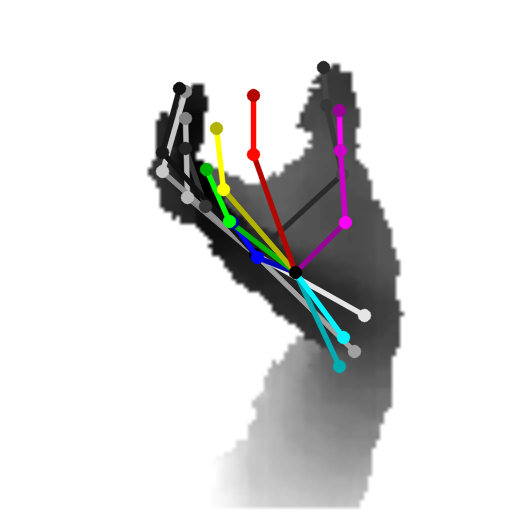} &
\includegraphics[width=\sampimgw,trim={3cm 2cm 2cm 1cm},clip]{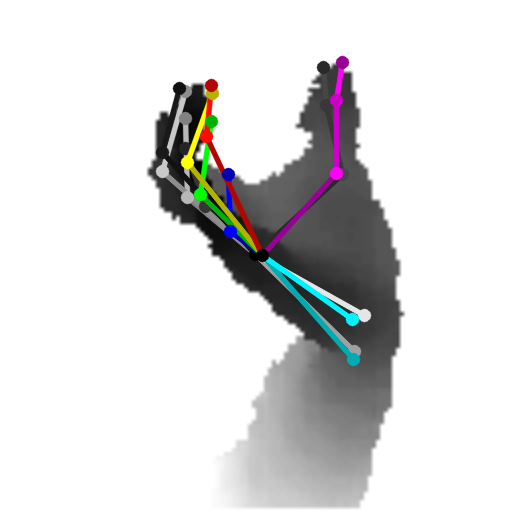} &
\includegraphics[width=\sampimgw,trim={3cm 3cm 1cm 4cm},clip]{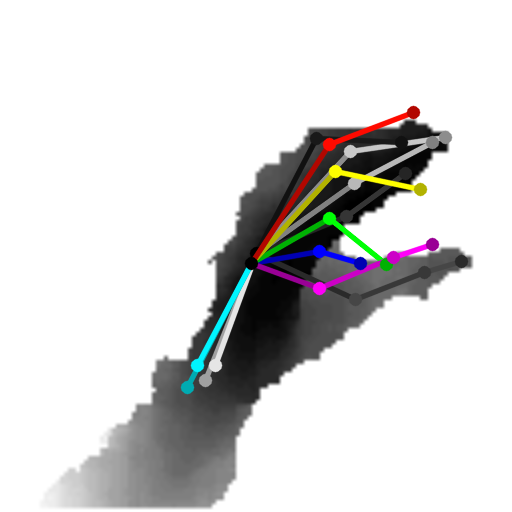} &
\includegraphics[width=\sampimgw,trim={3cm 3cm 1cm 4cm},clip]{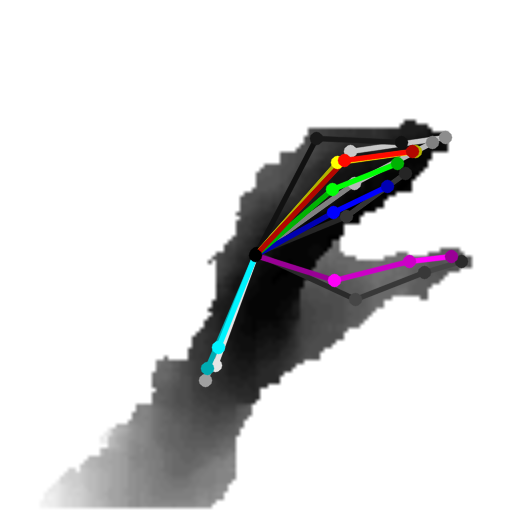} &
\includegraphics[width=\sampimgw,trim={3cm 1cm 2cm 1cm},clip,angle=90,origin=c]{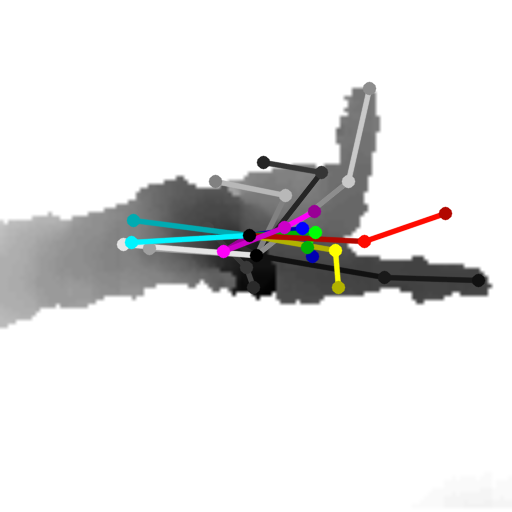} &
\includegraphics[width=\sampimgw,trim={3cm 1cm 2cm 1cm},clip,angle=90,origin=c]{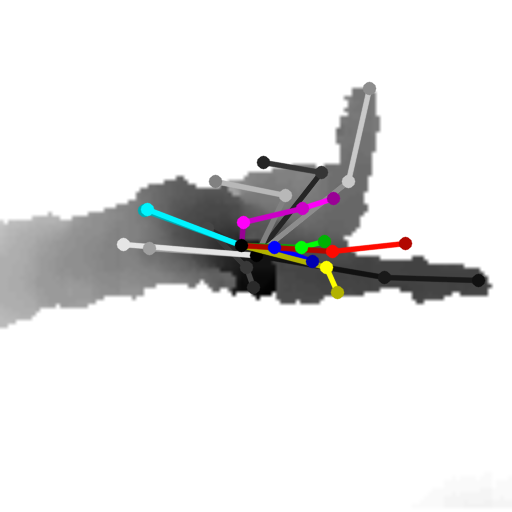}
\end{tabular}
\end{center}
   \caption{Qualitative results for NYU dataset. We show the inferred joint locations in color and the ground truth in grayscale projected to the depth images. The individual fingers are color coded, where the bones of each finger share the same color, but with a different hue. The left image of each pair shows the initialization and the right image shows the pose after applying our method. Our method applies to a wide variety of poses and is tolerant to noise, occlusions and missing depth values as shown in several images. The rightmost images show a failure case, where the error of the initialization is too large to recover the correct pose and the update only pushes the erroneous joints towards the hand silhouette.}
\label{fig:results_qualitative}
\end{figure*}
\egroup

Fig.~\ref{fig:results_qualitative}  shows some  qualitative examples.   For some
examples,  the predictor  provides  already  a good  pose,  which  we can  still
improve, especially for  the thumb.  For worse initializations, also larger updates on the pose can be achieved by our proposed method, to better explain the evidence in the image.

\section{Joint Hand-Object Pose Evaluation}
\label{sec:eval_ho}
In this section we present the evaluation of our approach for joint hand-object pose estimation on the challenging DexterHO dataset~\cite{Sridhar2016}. First, we describe the data we use for training. Then, we introduce the  benchmark  dataset.  Furthermore  we evaluate  our method qualitatively and quantitatively.

\subsection{Training Datasets}
Since there are no datasets for joint hand-object pose estimation available that contain enough samples to train a Neural Network, our approach relies on fusing real and synthetic data. We use real hand data and synthetic object data as described in Section~\ref{sec:training_data}. We use real hand data from the large MSRA~\cite{Sun2015} dataset, consisting of 76.5k depth frames of hands of 9 different subjects and a wide variety of hand poses. Further, we use the dataset of Qian~\etal~\cite{Qian2014}, which contains 2k depth frames of 6 different subjects. Both datasets were captured using a Creative RealSenz Time-of-Flight camera, which is the same camera as used for the experiments on the benchmark dataset. The depth image resolution is $320\times 240$px and the annotations contain $J=21$ joint locations.
The 3D object models are manually created from simple geometric primitives to resemble the objects from the benchmark dataset.

\subsection{Benchmark Dataset}
We evaluated our  method on the DexterHO dataset~\cite{Sridhar2016} for the task of joint hand-object pose estimation. The dataset contains several sequences of RGB-D data, totaling over 2k frames. There are two different objects used, a large and a small cuboid. The dataset has annotations for both the hand and the object available. The hand annotation contains the 3D location of visible fingertips, and the object annotation consists of three 3D points on the object corners. 
Although the dataset contains color and depth, we only use the depth images for the evaluation. During our experiments, we noticed some erroneous annotation that we fixed. We will make these corrected annotations available.

\subsection{Comparison with Baseline}
We evaluate the approach using the metric proposed by~\cite{Sridhar2016}, which combines the evaluation of the hand and the object poses:
\begin{equation}
E= \frac{1}{|V|+1_M} \left[\sum_{i\in V} \lVert X_i - G_i\rVert + \frac{1_M}{3}\sum_{m\in M} \lVert Y_m - F_m \rVert \right]
\end{equation}
The first sum measures the Euclidean distance between the predicted finger tips $X$ and the ground truth finger tips $G$ for all visible finger tips $V$. The second sum measures the Euclidean distance between the predicted object corners $Y$ and the ground truth corners $F$, where $M$ is the set of visible corners. Thereby the object pose is only evaluated if all corners of the object are visible,~\ie, the indicator function $1_M = 3$ if all three corners are visible.
Although we predict the 3D location of all joints of the hand, we only use the 3D locations of the finger tips for calculating the error metric. Also, we predict the full 6DoF pose of the object, and calculate the 3D corner locations for the evaluation.

We compare our results to the state-of-the-art on the DexterHO dataset~\cite{Sridhar2016} in Table~\ref{tab:dexho_quantitative}. Note that the baseline of~\cite{Sridhar2016} uses a tracking-based approach, which does not rely on training data but requires the pose of the previous frame as initialization. By contrast, we do not require any initialization at all and predict the poses from scratch on each frame independently. We outperform this strong baseline on two sequences (\textit{Rigid} and \textit{Occlusion}), and when~\cite{Sridhar2016} use only depth, as we do, we outperform their method on average over all sequences of the dataset. Our method is significantly more accurate for the object corner metric, since it is much easier to acquire training data for objects, compared to hands. For the accuracy of the hand pose estimation, our approach is mostly restricted by the limited training data.

The updaters perform two iterations for the hand and the object, since the results do not improve much for more iterations. Using the feedback loop improves the accuracy on all sequences by 10\% on average compared to our initialization.

Note that the two localizer CNNs take the same input data
  but predict different coordinates, since one is for the hand, the other one for
  the object. The average localization error of the hand gets 1\,mm
  worse when trained with spherical objects and tested on the DexterHO
  objects, \ie, cuboids. The localization error of the object gets 40\,mm worse when trained on spherical objects and tested on the DexterHO objects. This indicates that the hand localizer CNN, and similarly the hand pose predictor CNN, can still be used with other objects, but the object localizer CNN needs to be object specific.

We also evaluated an approach based on a single CNN
  predicting both hand and object 3D poses in order to assess the
  advantage of having two separate networks. In this case, we center
  the input of the CNN on the center-of-mass of the hand and the
  object combined together, as described in Section~\ref{sec:ho_loc},
  and the CNN outputs the poses of the hand and the object
  concatenated. Training this network appeared more  difficult and results in lower accuracy. The average combined error is 36.6\,mm, which is significantly more than our proposed approach using a separate CNN for the hand and the object, which achieves an error of 17.6\,mm.

\begin{table*}
\caption{Quantitative results on the DexterHO dataset~\cite{Sridhar2016}. Note that~\cite{Sridhar2016} uses color and depth information together with a tracking-based approach, which relies on a strong pose prior from the previous frame. In comparison, we perform hand and object pose estimation for each frame independently. Also, we use depth information only, for which their reported average error is larger than for our approach. Our approach significantly outperforms the baseline for the accuracy of the object pose on average.}
\label{tab:dexho_quantitative}
\centering
\begin{tabular}{c c | c c c c c c | c}
\toprule 
Method & Sequence & Rigid & Rotate & Occlusion & Grasp1 & Grasp2 & Pinch & Average \\
\midrule 
\multirow{3}{*}{\begin{tabular}{@{}c@{}}Sridhar~\etal~\cite{Sridhar2016}\\ RGB+Depth \end{tabular}} & Finger tips & 14.2\,mm & 16.3\,mm & 17.5\,mm & 18.1\,mm & 17.5\,mm & 10.3\,mm & 15.6\,mm \\ \cline{2-9}
 & Object corners & 13.5\,mm & 26.8\,mm & 11.9\,mm & 15.3\,mm & 15.7\,mm & 13.9\,mm & 16.2\,mm\\ \cline{2-9} 
 & Combined & 14.1\,mm & \textbf{18.0\,mm} & 16.4\,mm & \textbf{17.6\,mm} & \textbf{17.2\,mm} & \textbf{10.9\,mm} & \textbf{15.7\,mm} \\ \midrule

Sridhar~\etal~\cite{Sridhar2016} --- Depth only & Combined & -- & -- & -- & -- & -- & -- & 18.5\,mm  \\ \midrule

This work init --- Depth only & Combined & 14.1\,mm & 20.3\,mm & 18.8\,mm & 20.8\,mm & 24.3\,mm & 17.6\,mm & 19.3\,mm \\ \midrule

Single network --- Depth only & Combined & 29.0\,mm & 35.5\,mm & 35.8\,mm & 40.1\,mm & 39.7\,mm & 39.9\,mm & 36.6\,mm \\ \midrule

\multirow{3}{*}{\begin{tabular}{@{}c@{}}This work feedback\\Depth only\end{tabular}} & Finger tips & 14.2\,mm & 17.9\,mm & 16.3\,mm & 22.7\,mm & 24.0\,mm & 18.5\,mm & 18.9\,mm \\ \cline{2-9}
 & Object corners & 8.4\,mm & 23.4\,mm & 7.4\,mm & 8.2\,mm & 16.6\,mm & 9.6\,mm & 12.4\,mm \\ \cline{2-9}
 & Combined & \textbf{13.2\,mm} & 18.9\,mm & \textbf{14.5\,mm} & 20.2\,mm & 22.5\,mm & 16.7\,mm & 17.6\,mm \\ \bottomrule
\end{tabular}
\end{table*}

\subsection{Qualitative Results}
\begin{figure*}
\includegraphics[width=0.19\linewidth,trim={2cm 2cm 2cm 2cm},clip]{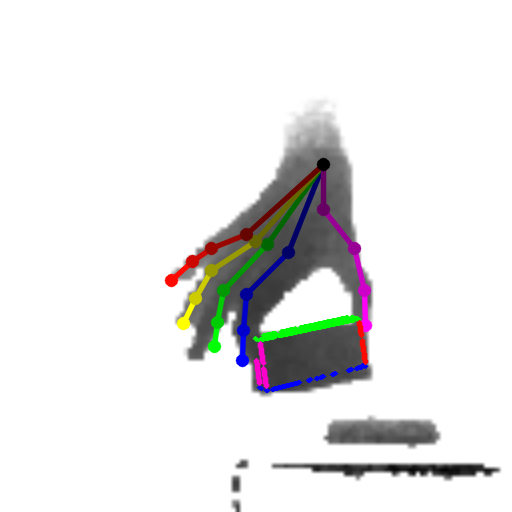}
\includegraphics[width=0.19\linewidth,trim={2cm 2cm 2cm 2cm},clip]{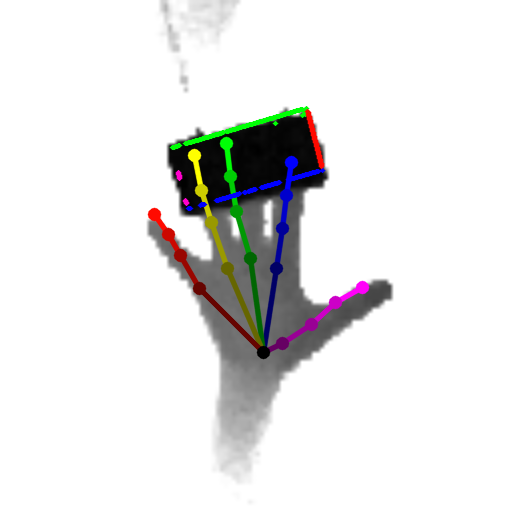}
\includegraphics[width=0.19\linewidth,trim={2cm 2cm 2cm 2cm},clip]{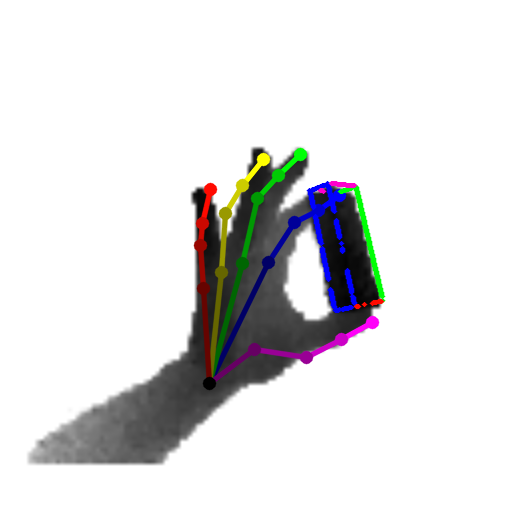}
\includegraphics[width=0.19\linewidth,trim={2cm 2cm 2cm 2cm},clip]{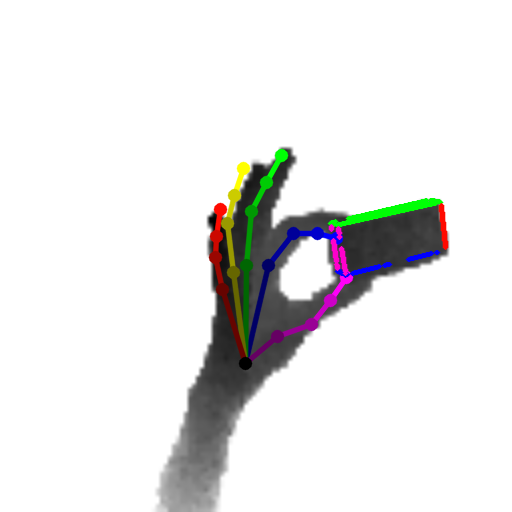}
\includegraphics[width=0.19\linewidth,trim={2cm 2cm 2cm 2cm},clip]{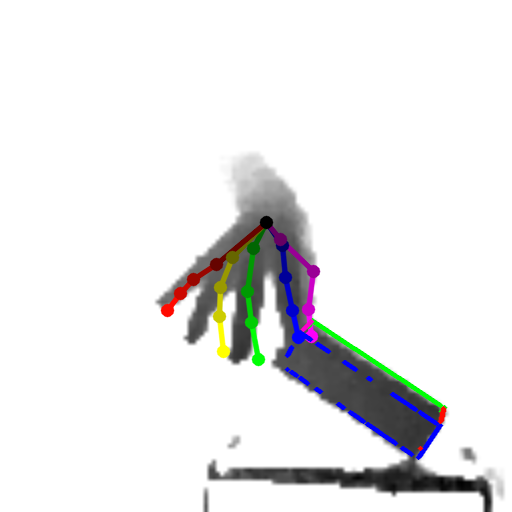}
\caption{Qualitative results on the DexterHO dataset~\cite{Sridhar2016}. Our method provides accurate hand and object pose.}
\label{fig:results_qualitative_dexho}
\end{figure*}

\begin{figure*}
\includegraphics[width=0.19\linewidth,trim={2cm 2cm 2cm 2cm},clip]{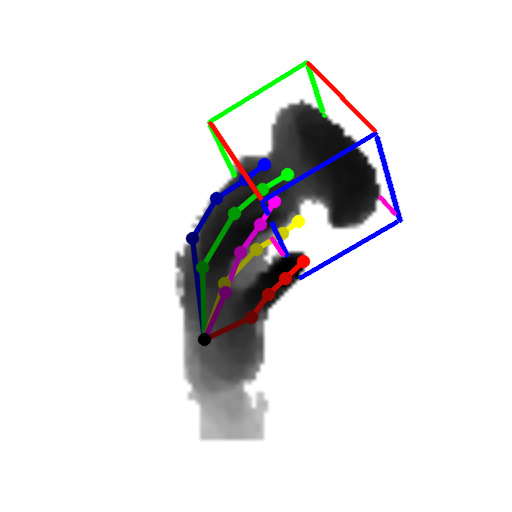}
\includegraphics[width=0.19\linewidth,trim={2cm 2cm 2cm 2cm},clip]{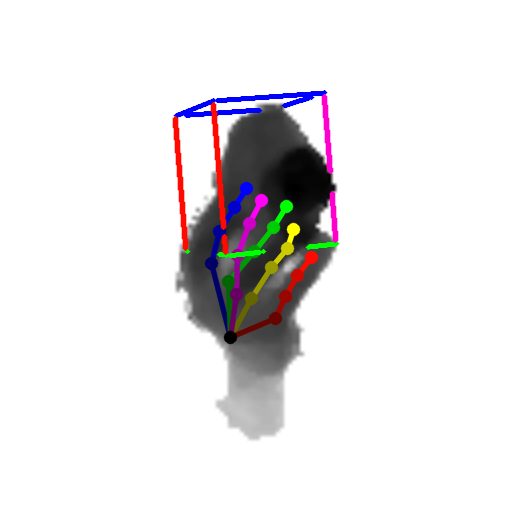}
\includegraphics[width=0.19\linewidth,trim={2cm 2cm 2cm 2cm},clip]{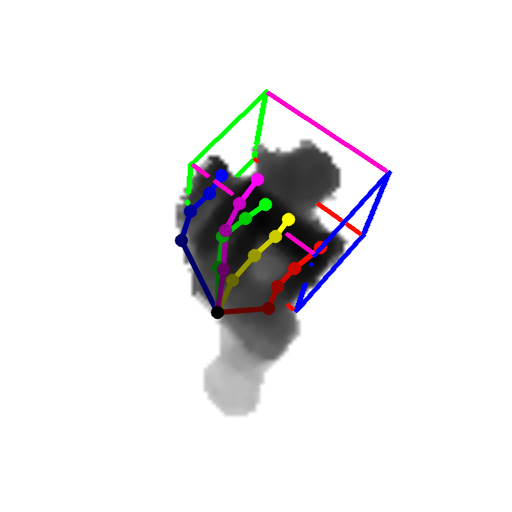}
\includegraphics[width=0.19\linewidth,trim={4cm 4cm 4cm 4cm},clip]{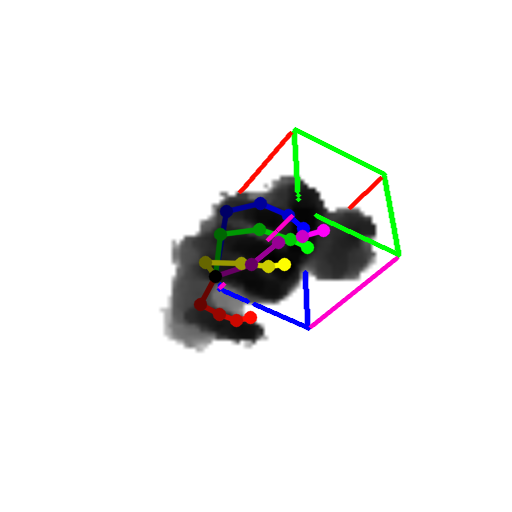}
\includegraphics[width=0.19\linewidth,trim={3cm 4cm 3cm 2cm},clip]{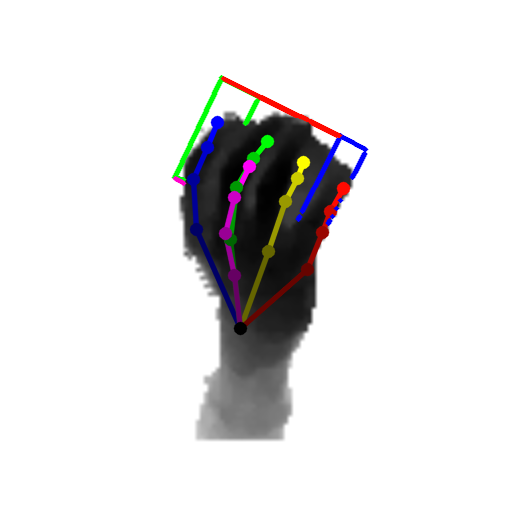}
\caption{Qualitative results for a hand interacting with a toy duck. Our method provides accurate hand and object poses, also when the hand is manipulating an object.}
\label{fig:results_qualitative_duck}
\end{figure*}

\begin{figure}
\begin{tabular}{lcc}
\rotatebox{90}{Initialization} & 
\includegraphics[width=0.4\linewidth,trim={2cm 2cm 2cm 2cm},clip]{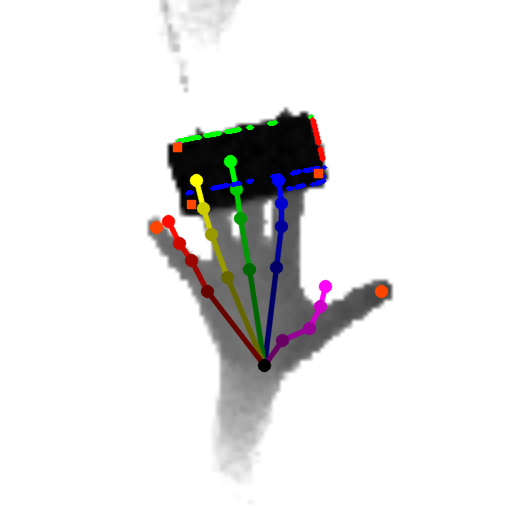} &
\includegraphics[width=0.4\linewidth,trim={2cm 2cm 2cm 2cm},clip]{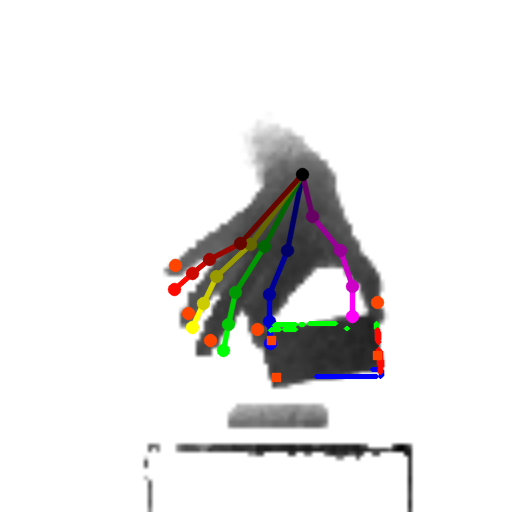} \\

\rotatebox{90}{Iteration 1} & 
\includegraphics[width=0.4\linewidth,trim={2cm 2cm 2cm 2cm},clip]{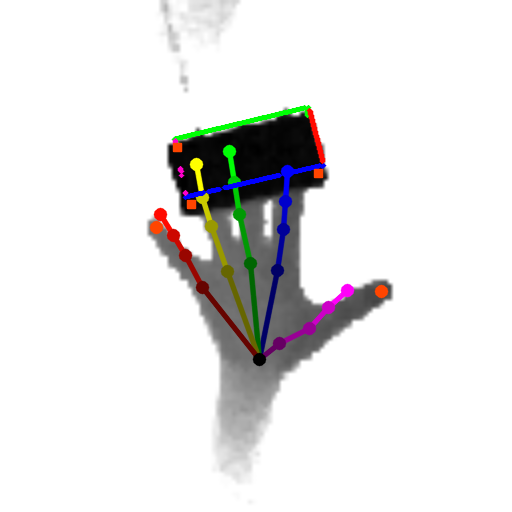} &
\includegraphics[width=0.4\linewidth,trim={2cm 2cm 2cm 2cm},clip]{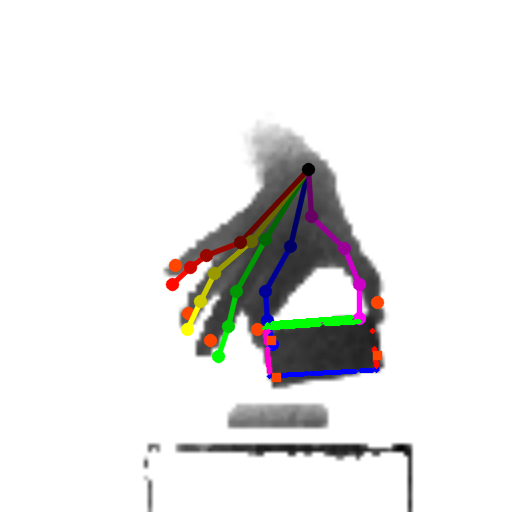} \\

\rotatebox{90}{Iteration 2} & 
\includegraphics[width=0.4\linewidth,trim={2cm 2cm 2cm 2cm},clip]{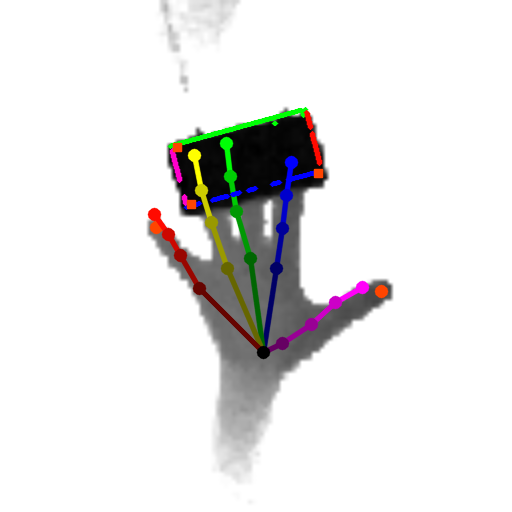} &
\includegraphics[width=0.4\linewidth,trim={2cm 2cm 2cm 2cm},clip]{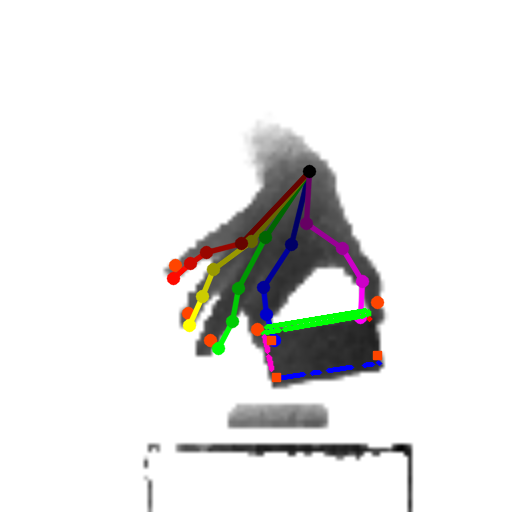} \\
\end{tabular}
\caption{Different iterations on the DexterHO dataset~\cite{Sridhar2016}. Top row shows the initialization, and further rows the consecutive iterations. Our results are shown in color, and the ground truth annotations are shown in orange. The ground truth contains the finger tips for the hand, and the corners for the object.}
\label{fig:dexho_iter}
\end{figure}

Fig.~\ref{fig:results_qualitative_dexho}  shows some  qualitative examples on the DexterHO dataset~\cite{Sridhar2016}, and Fig.~\ref{fig:results_qualitative_duck} shows qualitative results for a sequence where a hand is manipulating a toy duck. Our approach estimates accurate 3D poses for the hand and the object.\\
Fig.~\ref{fig:dexho_iter} shows the pose for consecutive iterations.  The predictor  provides  an initial estimate of the pose of the hand and the object, and  our feedback loop improves these initial poses iteratively.

\subsection{Runtime}

Our method is implemented in Python using the Theano library~\cite{Bergstra2010}
and we run the experiments on a computer equipped with an Intel Core i7, 64GB of
RAM, and an nVidia GeForce GTX 980 Ti GPU. Training takes about ten hours for
each CNN.
The runtime is composed of the localizer CNN that takes 0.8\,ms, the predictor CNN takes 20\,ms,
the updater  CNN takes  1.2\,ms for  each iteration, and  that already
includes the  synthesizer CNN with 0.8\,ms.  In  practice we iterate our  updater CNN twice, thus our method  runs at over 40\,fps on  a single GPU. 
For the joint hand-object pose estimation we have to run each network once for the hand and once for the object. We run the inference in two threads in parallel, one for the hand and one for the object, and use the predictor CNN with the simpler network architecture that takes 0.8~ms. Thus our approach runs at over 40\,fps on a single GPU.

\section{Conclusion and Future Work}
In this work we presented a novel approach for joint hand-object pose estimation. First, we separate the problem into hand pose and object pose estimation, in order to obtain an initial pose for the hand and the object independently. Then, we introduce a feedback loop, that refines these initial estimates. Remarkably, our approach does not require real data of hand-object interaction and can be trained on synthetic data, which simplifies the creation of the dataset. We evaluated our approach on public datasets. When considering hands only, our approach performs en-par with state-of-the-art approaches. For joint hand-object pose estimation our approach outperforms the state-of-the-art tracking-based approach when using only depth images.
Compared to such tracking-based approaches, our approach processes each frame independently, which is important for robustness to drift~\cite{Sharp2015,Rosales2006}.

It should be noted that our predictor and our synthesizer 
are trained with exactly the same data. One may then
ask how our approach can improve the first estimate made
by the predictor. The combination of the synthesizer and the
updater network provides us with the possibility for simply
yet considerably augmenting the training data to learn the
update of the pose: For a given input image, we can draw
arbitrary numbers of samples of poses through which the
updater is then trained to move closer to the ground truth.
In this way, we can explore regions of the pose space which
are not present in the training data, but might be returned by
the predictor when applied to unseen images.

This work can be extended in several ways. Given the recent trend in 3D hand pose estimation~\cite{Zimmermann2017,Mueller2018,Panteleris2018}, it would be interesting to adapt the feedback loop to color images, which means that the approach also needs to consider lighting and texture. Further, considering a generalization to an object class or different hand shapes would be interesting and could be achieved by adding a shape parameter to the synthesizer CNN. It would also be interesting to see how this approach works with a 3D hand CAD model instead of the synthesizer CNN. Future work could also consider the objective criterion of the updater training such that it would not require the hyperparameters for adding poses.


%

%

\ifCLASSOPTIONcompsoc
  \section*{Acknowledgments}
\else
  \section*{Acknowledgment}
\fi

The authors would like to thank Zeyuan Chen for his help with the
synthetic hand dataset. Prof.~Vincent Lepetit is a Senior Member of
the \emph{Institut Universitaire de France}.

\ifCLASSOPTIONcaptionsoff
  \newpage
\fi


\bibliographystyle{IEEEtran}
\bibliography{IEEEabrv,short,biblio}

%

\begin{IEEEbiography}[{\includegraphics[width=1in,height=1.25in,clip,keepaspectratio]{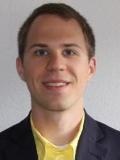}}]{Markus Oberweger}
received his MSc Degree in Telematics in 2014 and his PhD Degree in Computer Science in 2018, both from Graz University of Technology. He is currently a Research and Teaching Associate at the Institute for Computer Graphics and Vision at Graz University of Technology. His research interests include Machine Learning methods for Computer Vision, Deep Learning, and methods for 3D pose estimation.
\end{IEEEbiography}

\begin{IEEEbiography}[{\includegraphics[width=1in,height=1.25in,clip,keepaspectratio]{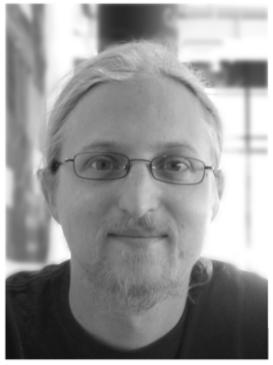}}]{Paul Wohlhart}
received his MSc Degree in Telematics in 2009 and his PhD Degree in 2014, both from the Graz University of Technology. During his PhD he was working on Machine Learning algorithms and Object Detection and Recognition. He then worked as a Post-doc at the Institute for Computer Graphics and Vision with Prof.~Lepetit. He now works as a research engineer at X, the moonshot factory. His research interests include Artificial Intelligence, Computer Vision, and Robotics.
\end{IEEEbiography}

\begin{IEEEbiography}[{\includegraphics[width=1in,height=1.25in,clip,keepaspectratio]{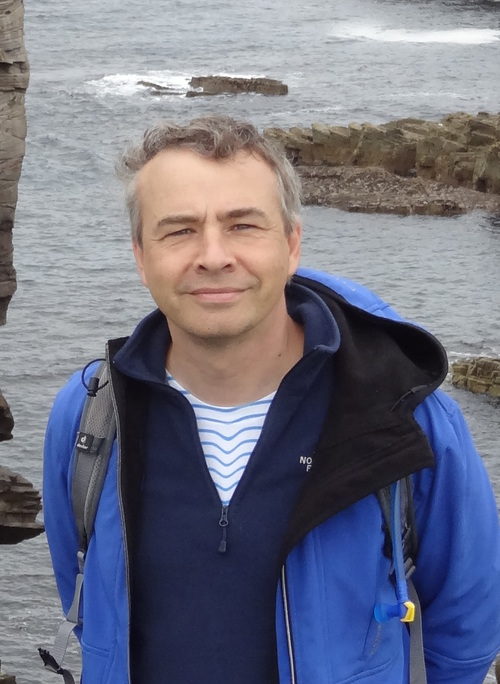}}]{Vincent Lepetit}
is a Full Professor at the LaBRI, University of Bordeaux, and a senior
member of the \emph{Institut Universitaire de France} (2018-2025).  He also
supervises a research group in Computer Vision for Augmented Reality
at the Institute for Computer Graphics and Vision, TU Graz. His
research interests include vision-based Augmented Reality, 3D camera
tracking,  Machine  Learning,  object recognition, and 3D
reconstruction. He is an editor  for  the IEEE Transactions on Pattern
Analysis and Machine Intelligence, the  International  Journal  of  Computer Vision, and the Computer Vision and Image Understanding journal.
\end{IEEEbiography}





\end{document}